\documentclass{article}

\usepackage{arxiv}

\usepackage[utf8]{inputenc} 
\usepackage[T1]{fontenc}    
\usepackage{hyperref}       
\usepackage{url}            
\usepackage{booktabs}       
\usepackage{amsfonts}       
\usepackage{nicefrac}       
\usepackage{microtype}      
\usepackage[title]{appendix}
\usepackage{graphicx}
\usepackage[bf]{caption}

\usepackage{pdflscape}      
\usepackage{booktabs}
\usepackage{longtable}
\usepackage{array}

\usepackage{doi}
\usepackage{etoolbox}
\usepackage[natbibapa]{apacite}
\usepackage{environ}

\AtBeginDocument{}
\renewcommand{\APACrefnote}[1]{}

\newtoggle{bibdoi}
\newtoggle{biburl}
\makeatletter
\newsavebox{\bib@url}
\newsavebox{\bib@doi}

\undef{\APACrefURL}
\undef{\endAPACrefURL}
\undef{\APACrefDOI}
\undef{\endAPACrefDOI}

\long\def\collect@url#1{\global\def\bib@url{#1}}
\long\def\collect@doi#1{\global\def\bib@doi{#1}}
\newenvironment{APACrefURL}{\global\toggletrue{biburl}\Collect@Body\collect@url}{\unskip\unskip}
\newenvironment{APACrefDOI}{\global\toggletrue{bibdoi}\Collect@Body\collect@doi}{}

\AtBeginEnvironment{thebibliography}{
 \pretocmd{\PrintBackRefs}{%
  \iftoggle{bibdoi}
    {\iftoggle{biburl}{\unskip\unskip \bib@doi}{}}
    {\iftoggle{biburl}{Retrieved from\bib@url}{}}
  \togglefalse{bibdoi}\togglefalse{biburl}%
  }{}{}
}

\makeatother

\title{Large-scale Building Damage Assessment using a Novel Hierarchical Transformer Architecture on Satellite Images}

\date{} 					

\renewcommand{\headeright}{}
\renewcommand{\undertitle}{}
\renewcommand{\shorttitle}{}

\hypersetup{
pdftitle={Large-scale Building Damage Assessment using a Novel Hierarchical Transformer Architecture on Satellite Images},
pdfsubject={},
}

\usepackage{color}
\usepackage{amsmath}
\definecolor{green}{rgb}{0, 0.5, 0}
\definecolor{orange}{rgb}{0.8, 0.6, 0.2}
\definecolor{red}{rgb}{1.0, 0.0, 0.0}
\definecolor{teal}{rgb}{0.0, 0.4, 0.4}
\definecolor{purple}{rgb}{0.65,0,0.65}
\definecolor{saffron}{rgb}{0.95,0.75,0.2}
\definecolor{turquoise}{rgb}{0.0,0.5,0.5}
\definecolor{black}{rgb}{0.0, 0.0, 0.0}
\definecolor{gray}{rgb}{0.5, 0.5, 0.5}

\newcommand{\dahitra}[0]{{DAHiTrA}}

\begin{document}
\maketitle

\begin{center}
{\Large
Navjot Kaur\textsuperscript{a},
Cheng-Chun Lee\textsuperscript{b},
Ali Mostafavi\textsuperscript{b,*}
Ali Mahdavi-Amiri\textsuperscript{a}, 
\par}

\bigskip
\textsuperscript{a} School of Computing Science, Simon Fraser University,\\ British Columbia, Canada\\
\vspace{6pt}
\textsuperscript{b} Urban Resilience.AI Lab, Zachry Department of Civil and Environmental Engineering,\\ Texas A\&M University, College Station, TX\\
\vspace{6pt}
\textsuperscript{*} correseponding author, email: amostafavi@civil.tamu.edu
\\
\end{center}
\bigskip
\begin{abstract}
This paper presents \dahitra, a novel deep-learning model with hierarchical transformers to classify building damages based on satellite images in the aftermath of natural disasters. Satellite imagery provides real-time and high-coverage information and offers opportunities to inform large-scale post-disaster building damage assessment, which is critical for rapid emergency response. In this work, a novel transformer-based network is proposed for assessing building damage. This network leverages hierarchical spatial features of multiple resolutions and captures temporal difference in the feature domain after applying a transformer encoder on the spatial features. The proposed network achieves state-of-the-art-performance when tested on a large-scale disaster damage dataset (xBD) for building localization and damage classification, as well as on LEVIR-CD dataset for change detection tasks. In addition, this work introduces a new high-resolution satellite imagery dataset, Ida-BD (related to the 2021 Hurricane Ida in Louisiana in 2021) for domain adaptation. Further, it demonstrates an approach of using this dataset by adapting the model with limited fine-tuning and hence applying the model to newly damaged areas with scarce data.
\end{abstract}


\section{INTRODUCTION}\label{sec1}
Rapid and automated damage assessment of buildings and infrastructure in the aftermath of disasters is critical to expedite emergency response and resources. Damage assessments done by ground crews can be time-consuming and labor-intensive \citep{spencer_advances_2019}. The number of studies focused on damage assessments for civil infrastructures and buildings has grown recently. Several studies are dedicated to identify damage and monitor structure health \citep{amezquita-sanchezNovelMethodologyModal2017, amezquita-sanchezWirelessSmartSensors2018, liNewMethodModal2017,ohEvolutionaryLearningBased2017,jrInfraredThermographyDetecting2018,perez-ramirezRecurrentNeuralNetwork2019,wuPruningDeepConvolutional2019,xuRealtimeRegionalSeismic2021}. In addition, recent studies demonstrate the capability of damage assessments on civil infrastructures and buildings with computer-aided approaches, including damage assessment for buildings \citep{zhouAutomatedAnalysisMobile2018,sajediVibrationbasedSemanticDamage2020,gaoDeepTransferLearning2018}, bridges \citep{talebinejadNumericalEvaluationVibrationBased2011,zhangConcreteBridgeSurface2020}, and concrete structures \citep{zou_multicategory_2022, athanasiouMachineLearningApproach2020,liAutomaticPixellevelMultiple2019}. Due to the advancement of remote-sensing, imagery can usually be obtained within a few days after disaster events (e.g., WorldView-2 satellite has a revisit frequency of 3.7 days or less at 20 degrees off-nadir \citep{WorldView2}) and thus provide opportunities for assessing damage conditions more efficiently at a larger scale than manual visual inspection. To investigate expediting building damage assessments, studies have applied computer-vision techniques on building images \citep{wang_autonomous_2020,zou_multicategory_2022}, high-resolution aerial imagery \citep{fujita_damage_2017,cheng_deep_2021, khajwalPostdisasterDamageClassification2022}, and satellite imagery \citep{mccarthy_mapping_2020,tong_building-damage_2012,cao_building_2020}. While aerial imagery can capture more details about buildings' conditions due to lower flying altitudes than satellite imagery, it requires extensive local planning and provides a smaller area of building coverage than satellite imagery. Satellite imagery, which provides near real-time and high-coverage information, offers opportunities to assist in large-scale post-disaster building damage assessments \citep{corbane_comparison_2011}. Studies \citep{siam-att-unet2, siam-att-unet, bdanet} have shown that by leveraging satellite imagery and deep learning, the process of damage assessments can be accelerated with the generation of high-quality building footprints and damage-level classification for each building. The majority of existing models are built using a single dataset (xBD) \citep{xBDdataset} but have not been tested on datasets that include newly damaged areas. This work overcomes these limitations by creating a novel deep-learning technique to perform damage classification, building segmentation tasks, and testing of the model on a new and higher-resolution dataset related to building damage, Ida-BD, in Louisiana in the aftermath of Hurricane Ida. 

The most common approach for building damage assessment using satellite imagery is to pose the problem as a combination of segmentation and classification tasks and to train deep-learning models on pre-disaster and post-disaster satellite images. Many researchers have utilized convolutional neural networks (CNNs) to achieve models with acceptable performance \citep{siam-att-unet2, siam-att-unet, rescuenet}. For example, \citet{detectron} considered building damage assessment as a semantic segmentation task in which damage levels are assigned to different class labels. Pre- and post-disaster image channels are concatenated into a single input, relying on the model to differentiate between the channels of pre- and post-disaster conditions, which complicates the learning process. The problem of damage classification can be also modeled as change detection where the change is detected over a period of time with a pair of pre- and post-disaster images. Standard practice in the existing literature is to use a CNN-based network with a two-branch architecture for pre- and post-disaster images and directly concatenating the learned features to process pixel localization and classification \citet{siam-att-unet2,siam-att-unet,rescuenet}. 

The proposed method, building damage assessment using a hierarchical transformer architecture (DAHiTrA), argues that directly concatenating features complicates the task of localizing pixels for damage assessment; the network should concentrate on the changes between pre- and post-disaster images. Therefore, the proposed classifier works on the \emph{difference} between features. Since pre- and post-disaster images are usually obtained at different times and under different lighting or weather conditions, to find meaningful and unbiased difference, both images should be mapped to a \emph{common domain}. Therefore, the proposed method used \emph{difference blocks} that map the features of post- and pre-disaster images onto a common domain to better assess the difference. In addition, as damage is typically at different scales, features should be obtained at multi-resolutions. Therefore, the proposed network takes advantage of UNet-based structures to learn features at different scales and build a \emph{hierarchy} of those features obtained at different scales to localize and classify the changes between two images. In Section \ref{sec5}, ablation studies are performed to validate the proposed design choices, to compare the proposed method with the state-of-the-art techniques, and to demonstrate that it outperforms the alternatives.

Every disaster is unique although they may share some similar characteristics. For example, the building damage caused by a hurricane may not be the same as the damage caused by another hurricane in the future. Therefore, using models based on images directly from previous disaster events to assess the damage of a new event requires the models to be tested on new datasets to evaluate the adaptation performance of the models. To address this domain shift from new disaster events, one approach could be to train the model on a subset of disaster events from xBD dataset and then test on other events. Another challenge with satellite imagery for damaged areas is the possibility of them being assessing from different cameras (for instance, a different satellite) and hence it could have different resolution or other properties. Further, the satellite imagery providers may apply different pre-processing methods. Hence, the approach of training and testing on the imagery from the same source does not handle this challenge. This has been a major limitation in previous studies since most of the existing models were created based on the xBD dataset with limited or no adaptation performance testing. To address this limitation, a new dataset, Ida-BD, is introduced with 87 image pairs (1,024 $\times$ 1,024) with a very high resolution (0.5 m/pixel) from Hurricane Ida (2021) in Louisiana. Hurricane Ida brought the strongest winds ever recorded in Louisiana and heavy rains, which caused an estimated \$18 billion in building damage and at least 30 fatalities in Louisiana \citep{beven_ii_national_2022}. Since the near-real-time damage assessment face the scarcity of labeled datasets, already available datasets can be used for training, and some domain adaptation techniques or fine-tuning can be applied to obtain satisfactory results on the newly damaged areas with scarce data. As a result, Ida-BD dataset can serve as a benchmark for domain adaptation from larger datasets like xBD. A baseline is provided for the domain adaptation task using a simple transfer learning method.
Accordingly, the contributions of this work are as follows:
\begin{itemize}
\item A hierarchical UNet-based architecture is presented, which uses transformer-based difference to perform well on both the damage classification and building segmentation tasks. \textbf{This is the first method that explicitly relies on the difference of transformer-encoded features from pre-disaster and post-disaster images and hierarchically builds the output from multi-resolution features}.
\item A new dataset, Ida-BD, is introduced for evaluating the model's adaptation performance and provide a baseline for damage assessment task. It also demonstrates the usefulness of the new dataset and efficacy of the proposed network using transfer learning.
\end{itemize}

The rest of the paper is organized as follows. The related works are first discusses in the field of building damage classification and change detection using satellite images in Section \ref{sec2}. In Section \ref{sec3}, the proposed model architecture along with loss functions and training settings are discussed. Section \ref{sec4} provides an overview of the datasets used for evaluation and the results are presented in the Section \ref{sec5}. Section \ref{sec6} concludes the paper.
\section{Related work}\label{sec2}

\begin{figure*}[t]
\centerline{\includegraphics[width=0.95\linewidth]{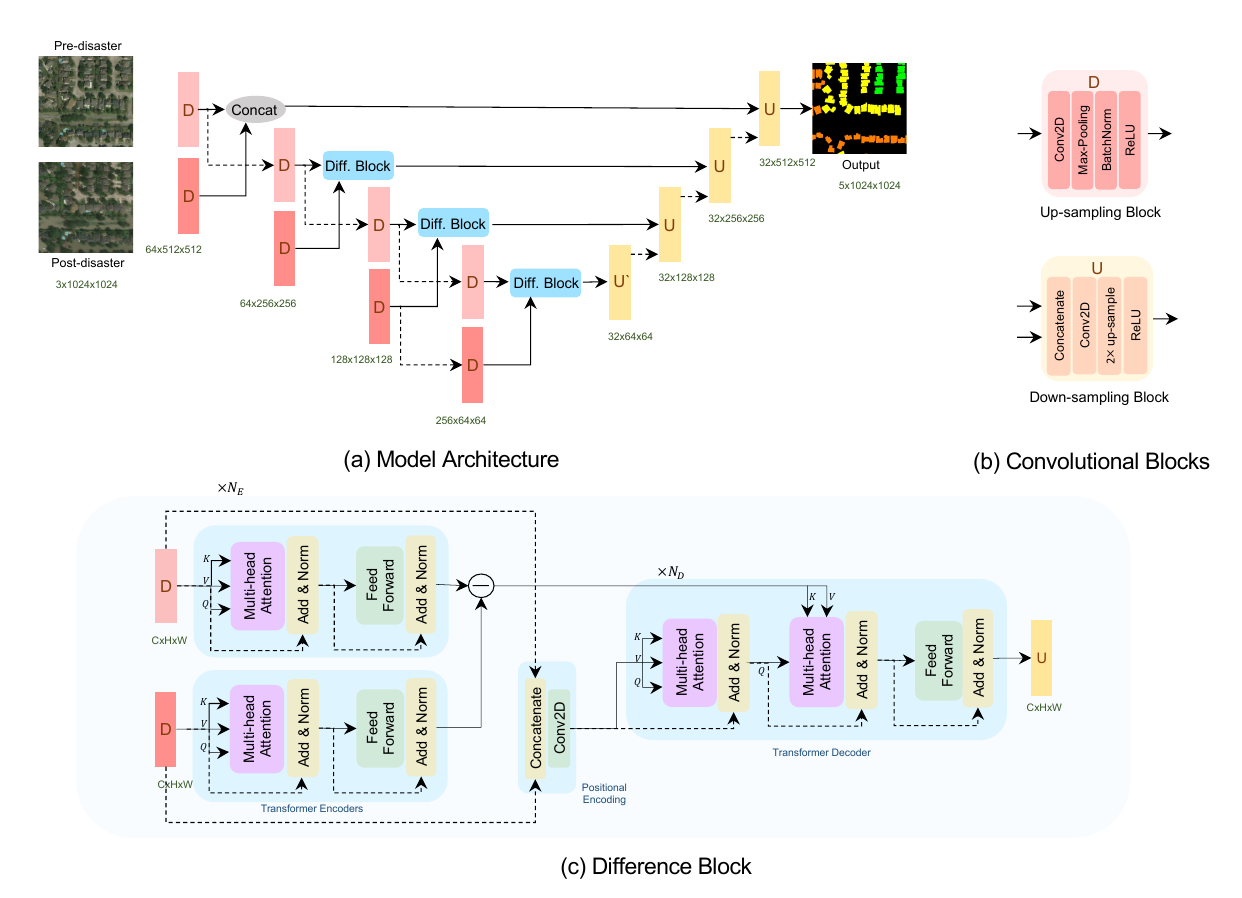}}
\caption{The model architecture for damage detection and classification. A pair of pre-disaster and post-disaster images is fed into an encoder made up of stacked convolution block as (red layers), where each convolutional block downsamples the features space (dashed lines). The output from each block of pre- and post- disaster image encoders is passed jointly (solid lines) into the difference block, which maps the spatial features into feature domain and then take the difference followed by transformer decoder to map the features back into spatial domain. Next, the output mask is hierarchically built by up-sampling (dashed lines) and concatenating the features from lower dimension to higher dimension layers. The first up-sampling block is an exception which takes in single input instead of two inputs.}
\label{ProposedArchitecture}
\end{figure*}

This section discusses recent methods used for building damage classification using satellite images. Since the xView2 challenge \citep{xBDdataset} in 2019, many researchers have tried various techniques from Siamese networks \citep{siam-att-unet, siam-att-unet2}, to convolutional networks \citep{detectron, rescuenet}, to attention mechanisms \citep{bdanet} for solving the problem of damage classification.

Earlier models, such as the one done by \citet{siam-att-unet2}, combined the UNet model with the Siamese architecture where the UNet model learns the semantic segmentation of buildings and the Siamese network focuses on the damage classification by comparing the segmented outputs. \citet{siam-att-unet} further developed the model by using attention-based UNet where attention is applied to the incoming layers from the encoder module before fusing with the up-sampled features in the decoder. These architectures rely on global features learned from pre- and post-disaster images that are then merged to learn the change (here, damage) at the final stage. Since the majority of parameters prioritize learning the global features for a single image, this setting makes it difficult for the model to learn the change from the benefit of the multi-task supervision.

\citet{rescuenet} drew insights from DeepLabv3+ and came up with RescueNet model. They used a backbone of Dilated ResNet and Atrous Spatial Pyramid Pooling module to generate multi-scale features from pre- and post-disaster images and then compared their differences to learn the temporal change. Similar to the concern with the Siamese networks, a large set of layers of this model focuses on learning the changes in images separately, which limits the learning capacity of the last few layers to make an efficient change detection. Additionally, this model uses multi-scale features to capture input image variations, but these features are later simply concatenated and passed into a convolutional block for the final output. Because the change itself could be of varying resolutions, a hierarchical or multi-scale analysis would be useful to generate a better quality temporal change mask. 

More recently, researchers modeled the temporal relation between pre- and post-disaster images using attention mechanisms. For instance, \citet{bdanet} introduced BDANet where they use cross-directional attention in a two-stage framework for building segmentation and damage assessment. After modeling for building segmentation using single UNet branch, the pre- and post-disaster images are fed into two separate branches, and cross-directional attention is used to exchange information between the two parameter branches. It is noted that rather than just making the information available from the other temporal domain, one can explicitly enforce the model to learn the temporal difference by taking their absolute difference in a common feature domain. 

To support temporal difference in common domain, the self-attention can be used to map the features from spatial to the feature domain, which can be efficiently applied through transformers. The transformers, although not specifically used for building damage classification tasks, have been successfully applied for building change detection on satellite images. It can be seen that the damage classification problem (multi-class classification) is similar to \textit{change detection} (binary classification) by viewing the level of damage as change, although the damage classification problem is more difficult due to the higher target dimensionality and higher inter-class skewness. Moreover, the damage classes are not mutually exclusive and distinct as compared to change detection where the presence of new infrastructure directly indicates change. For instance, there is no definitive way to classify the roof damage as major damage instead of minor damage.

 \begin{figure*}[t]
\centerline{\includegraphics[width=\linewidth]{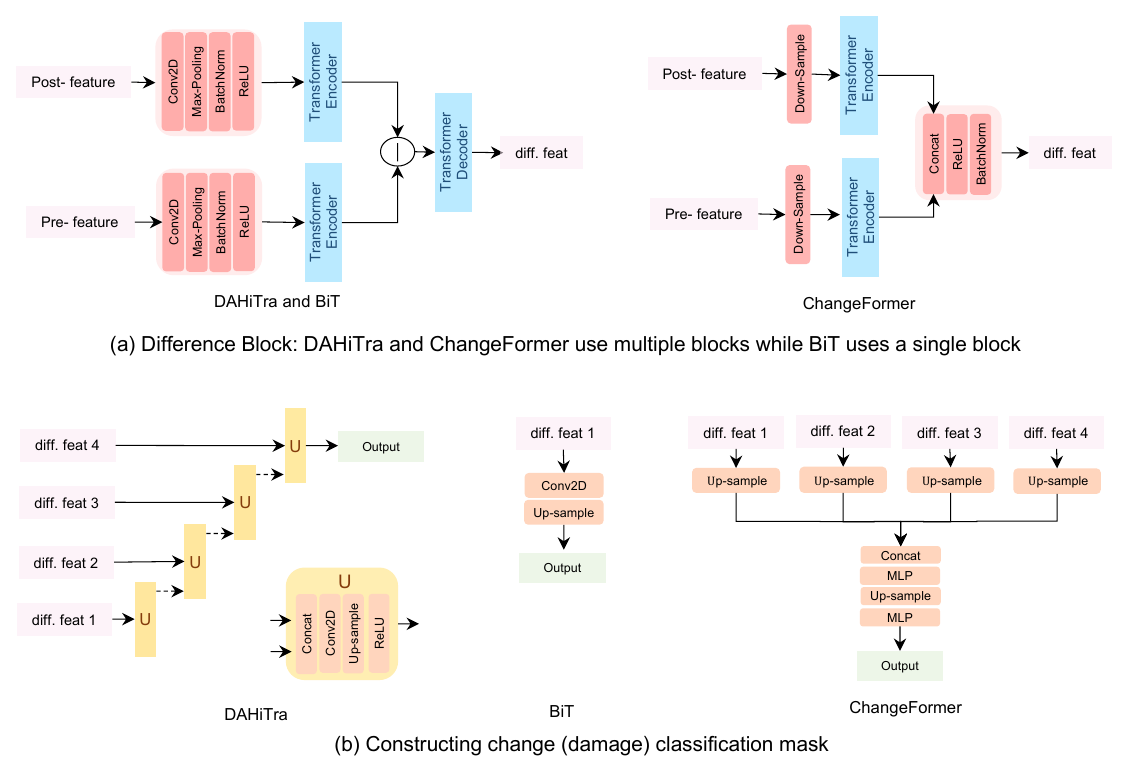}}
\caption{Comparing the model architecture of \dahitra{} (ours) with two recent works for change detection, ChangeFormer \citep{changeformer} and BiT \citep{chen2021bit}, for reference. a) Difference blocks: \dahitra{} and BiT use transformer encoders to map pre- and post- change features into a common domain, take the absolute difference, and then use a transformer decoder to map back the features in the spatial domain. \dahitra{} uses 4 such blocks to get difference features of varying resolutions while BiT only uses a single resolution. In ChangeFormer, the pair of images are passed into transformer encoders and then output features are concatenated to get difference features. b) Constructing the outputs: \dahitra{} construct hierarchical output from the four difference features; BiT upsamples the single difference feature to construct the output; ChangeFormer concatenated four difference features and pass them into multi-layer perceptron stack to get the final output.}
\label{RelatedModels}
\end{figure*}

The state-of-the-art methods for the change detection problem include transformer-based models, as discussed following. \citet{chen2021bit} introduced bi-temporal image transformer (BiT) \ref{RelatedModels} to model pre- and post-change contexts into visual semantic tokens using spatial attention. The transformer encoder is used to learn the temporal context between the two token sets, followed by a transformer decoder projecting the features back to the pixel space. An explicit difference is then taken on the features to generate the change mask. Although this model has significantly outperformed previous models for change detection, its performance is limited by the feature space of the visual semantic tokens. The tokens are a pair of low-dimension features generated by a CNN module, which are passed into a single transformer encoder and decoder to learn change and reconstruct a high-dimensional output. Also, the final change prediction mask is generated by a single step up-sampling from the very low-resolution feature space (1/4th of the final resolution) which makes it difficult to generate high-resolution change masks. In DAHiTrA's architecture, the model considers multi-resolution features which pass into different transformer encoders and decoders to learn change at different resolutions, and then instead of single-step upsampling, hierarchical upsampling is used on low and high-dimensional features to construct the final output.. 
 
Another recent work for change detection is ChangeFormer by \citet{changeformer}, which uses hierarchically structured transformer encoder modules along with multi-layer perceptron (MLP) decoders in a Siamese network architecture \ref{RelatedModels}. In ChangeFormer, transformers function as feature extractors and are applied on all scales of features; however, as analyzed by \citet{Wu_2021_ICCV}, the transformers are inefficient at the early stages, and the convolutions are better at extracting low-level features. Additionally, the number of parameters for transformers increases four times with the size of the image which makes the former difficult to use for large-size images as input. Therefore, our work instead uses the convolutions as the base feature extractors and applies the transformers to map these features from different temporal domains to a common domain and then projects them back using transformer decoders. Secondly, in the work by \citet{changeformer}, the output mask is based on parallel concatenation of different resolution change masks, as outputted from the transformer encoder and then passed into a Multi-layer perceptron to generate the final output. However, in \dahitra{}, the output is constructed using pairs of transformer encoder and decoder, and the change is learned explicitly in the transformer encoded domain which is mapped back into the spatial domain. The output is then constructed hierarchically from low-resolution to high-resolution features which provide the model more guidance for global context and joint optimization of multi-resolutions and hence produce better high-resolution results. As discussed in Section 5.2, \dahitra{} outperforms ChangeFormer and other recent works for the damage classification and change detection tasks.

Few other transformer based change detection works include MSCANet \citep{cropland} with CNN-transformer hybrid architecture with multi-scale context aggregation for cropland change detection and EGCTNet \citep{egctnet} with a fusion encoder that combines convolutional network with transformer features. Although these networks use transformer encoder for feature encoding, they do not explicitly enforce the network to learn the temporal difference. This work is the first method that explicitly relies on the difference of transformer-encoded features from pre-disaster and post-disaster images and hierarchically constructs the output from multi-resolution features.
\section{Method}\label{sec3}

This work proposes a novel neural network called \dahitra{} for multi-class change detection for the purpose of building damage assessment which is based on UNet and transformers to learn hierarchical features of pre- and post-disaster images. The multi-resolution properties of UNet are utilized to establish a hierarchical feature based difference block that can achieve highly accurate classification and segmentation results. This is the first method that explicitly relies on the differences of transformer-encoded multi-resolution features from pre-disaster and post-disaster images and then hierarchically builds the output from these multi-resolution difference features (Figure \ref{ProposedArchitecture}). Here, \dahitra's architecture and loss functions are explained. To show that \dahitra can be used for unseen small datasets, a domain adaptation techniques has been discussed.

\subsection{Motivation}
\dahitra{} is motivated by the strengths of visual transformer and UNet architecture, which are established for effectively learning context and image segmentation tasks, respectively. The UNet \citep{unet} is a powerful fully convolutional architecture where the usual contracting network is supplemented by successive layers by replacing pooling operations with upsampling operators. The contracting layers of the UNet model are made up of convolutional blocks, which is a stack of convolution, batch normalization, non-linear activation (ReLU) and max-pooling (scale of 0.5). The expanding layers of the UNet are made up of similar convolutional blocks but without the max-pooling. Instead, upsampling (scale of 2) is used to double the size of the spatial features at every layer. Hence, for an input of 1,024 $\times$ 1,024 image and model with 4 contracting and 4 expanding layers, the spatial size reduces to 64 $\times$ 64 at the bottleneck layer, which is then up-sampled back to generate output with the same size as input. Since the UNet computes pixel-wise output and works well with a small amount of data, it is a simple yet well fitted network for semantic segmentation problems. 

On the other hand, visual transformers \citep{Wu_2021_ICCV} extract patches from images and feed them into transformer encoders to learn relationships between different image regions and hence obtain a global representation. The transformer encoders are made up of multi-head self-attention layers (MSA). The basic idea of MSA is to apply multiple independent attention heads in parallel and then concatenate and project the output to the required mapping size through a multi-layer perceptron. There are three inputs to an attention head: key $K$, query $Q$ and value $V$, which are computed from the input $X$ $\in R^{L \times C}$ ($L$: token length; $C$: channels) as following: 
\begin{align*}
    K &= X W_k \\
    Q &= X W_q \\
    V &= X W_v
\end{align*}
where $W_k, W_q$ and $W_v \in R^{C\times d}$ and are trainable parameters. A single attention head is formulated as: 
\begin{equation}
   \text{Att}(K, Q, V) = \sigma \left(\frac{QK^T}{\sqrt{d}}\right)V
\end{equation}
The multi-head self-attention concatenates multiple such attention heads:
\begin{equation}
   \text{MSA}(X) = \text{Concat}(\text{head}_1,...,\text{head}_n)W_o
\end{equation}
where $\text{head}_j$ = Att($K_j,Q_j,V_j$), $n$ is number of attention heads and $W_o \in R^{nd \times C}$ is the linear projection matrix for output. To project the output back into the spatial domain of image, transformer decoders are used. Similar to the transformer encoders, the decoder modules are also based on multi-head self-attention layers which take the features from the transformer encoder as an input. The transformers are well suited for learning complex image embeddings due to their high modeling capacity, global receptive fields and low inductive bias when compared to CNNs. 

Using the insight from \citet{Wu_2021_ICCV}, the transformers are inefficient at the early stages and the convolutions are better at extracting low-level features. In this work, we combine the respective strengths of UNet and transformers by using the convolutions as the base feature extractors and then using the transformers to learn the temporal difference by mapping the features into a common domain. Hence, this work combines the respective strengths of UNet and transformers as we discuss below.

\subsection{Model pipeline}
Initially a UNet model is trained for building segmentation task on the xBD dataset using pre-disaster images while intermittently giving few post-disaster images. This model serves as a pre-trained backbone for the damage assessment task. In order to learn the difference between pre-disaster and post-disaster images, the UNet network is split into the encoder \begin{math}E\end{math}, which is the first half with contracting levels and decoder \begin{math}D\end{math}, which is the second half with the expanding levels. The encoder \begin{math}E\end{math} is replicated with shared weights and the pre- and post- images are independently fed into \begin{math}E_{pre}\end{math} and \begin{math}E_{post}\end{math}. Then the features from the \(ith\) contracting level \begin{math}E_{x}(i)\end{math}, where \(x \in \{pre, post\}\), is passed into the \textit{difference block} \begin{math}Z\end{math} \ref{3.2.1}, which gives the joint difference features \begin{math}{z}_{i}\end{math} as output. The number of contracting levels is set to four. Since each contracting level is of different resolution, we get multi-resolution difference features.
\begin{align}
\mathcal{z}_{i} &= \mathcal{Z}_{i}(E_{pre}(i), E_{post}(i)) & i \in \{1..4\}
\end{align}

Next, the decoder \(D\) is used to reconstruct the pixel-wise output. The difference feature \begin{math}{z}_{i}\end{math} is passed to the \(ith\) expanding level of the decoder \begin{math}D\end{math} which then hierarchically builds the output mask using up-sampling blocks. {Each up-sampling block takes in the output feature from the difference block and the previous up-sampling block. Both the features are concatenated channel-wise and hence increasing the channel dimensions. The concatenation is followed by convolutional layers to avoid any artifacts due to up-sampling as well as to reduce the channel dimensionality. Next, the feature is up-sampled in spatial space to get an output feature of higher spatial dimension. This is repeated to finally get the output feature of the desired size, i.e the size of the input imagery. This leads to hierarchical development of the final damage output mask of $n$ channels where $n$ is a total number of classes and each channel contains the probability of the corresponding class.} While training, a softmax function is applied to get an output whose values lie between $[0-1]$ and sum up to 1.
\begin{align}
    \text{Softmax}(x_i) = \frac{\exp(x_i)}{\sum_j\exp(x_j)}
\end{align}
While evaluation, 'argmax' is applied on the final layer of $c$ channels (instead of 'softmax') to get a single channel mask whose each pixel $p_i$ contains the index for the most probable class $c_i$. The 'argmax' function simply returns the channel index $i$ for which the probability confidence is maximum. For instance, for an input of a pair of 3 $\times$ 1,024 $\times$ 1,024 images and 5-class classification model, the output is 1,024 $\times$ 1,024 mask, where each pixel contain 0-4 values. Here, 0 represents the background and 1-4 represents the buildings with no-damage to completely destroyed state.

\subsubsection{Difference block}\label{3.2.1}
An important component of \dahitra{} is the \textit{difference block} which is made up of transformer modules. The difference block at level $i$ takes in features $E_{x}(i)$ where $x \in \{pre, post\}$ from the $ith$ contracting level of UNet encoders $E_{x}$ with images from the two different temporal domains as inputs (pre-disaster and post-disaster). The features are independently passed into a transformer encoder $T_{ei}$ with shared weights and hence encoding the features from different temporal domain to a common feature domain. An absolute difference of the encoded features is calculated and then the difference is mapped back into the original spatial domain using transformer decoder $T_{di}$.
\begin{align}
{Z}_{i}(E_{pre}(i), E_{post}(i)) & = T_{di}(K) \\
\text{where } K & = \|T_{ei}(E_{pre}(i)) - T_{ei}(E_{post}(i))\|\nonumber
\end{align}

\subsection{Loss function} This work used a class-weighted loss function provided in Equation \ref{loss_functions} with a combination of focal loss \citep{focal_loss} and dice loss \citep{dice_loss} to perform well for the classification task on the unbalanced dataset. The focal loss $\mathcal{L}_{foc(i)}$ is used to further address the class imbalance where a modulating term is applied to the cross-entropy loss function to focus the learning on hard mis-classified examples. The second loss is dice loss $\mathcal{L}_{dice(i)}$, which tries to maximize the overlap between predicted and ground-truth boundaries. Its denominator considers the total number of boundary pixels at global scale, while the numerator considers the intersection, implicitly capturing local behaviour. An $\alpha$ hyperparameter is used to balance the impact of focal loss and the dice loss. The class weights $w_i$ are assigned as per the pixel distribution of the classes in the dataset.
Hence for $n$ classes, the loss function is as following:
\begin{align}\label{loss_functions}
\mathcal{L}_{dmg} &= \sum_i w_i (\mathcal{L}_{foc(i)} + \alpha \mathcal{L}_{dice(i)}) & i \in \{0,\dots n\}\\\nonumber
\mathcal{L}_{foc(i)} & = \sum_i - (1 - p_i)^\gamma \log(p_i) & i \in \{0,1\}\\ \nonumber
\mathcal{L}_{dice(i)} & = 1 - \frac{2 \mid \tilde{c} \cup c \mid}{\mid \tilde{c}\mid \cap \mid c\mid } & c \in \{0,1\} 
\end{align}
For instance, in the xBD dataset discussed in Section \ref{sec4.1}, there are five classes: background (0), no damage (1), minor damage (2), major damage (3) and destroyed (4), and weights $w_i$ were assigned as per the pixel distribution of the classes in xBD dataset (Table \ref{tab_xbd_classweight}). On the other hand, the building change detection problem is a binary classification problem and we do not use additional weights in the loss function and rely on the focal loss to account for the class imbalance. Hence the loss function for the building change detection problem is a combination of focal loss and dice loss as following:
\begin{align}
\mathcal{L}_{chg} &= \mathcal{L}_{foc(i)} + \alpha \mathcal{L}_{dice(i)} & i \in \{0,1\}
\end{align}

\begin{table}[!htbp]
\centering
\caption{Class-wise pixels count distribution in xBD dataset and the corresponding weights used in training. The notation used: 0-Background, 1-No damage, 2-Minor damage, 3-Major damage, 4-Destroyed}\label{tab_xbd_classweight}
\begin{tabular*}{\columnwidth}{@{\extracolsep\fill}lccccc@{\extracolsep\fill}}%
\toprule
\textbf{Class Id} & \textbf{0} & \textbf{1} & \textbf{2} & \textbf{3} & \textbf{4} \\
\midrule
Pixels \% & 96  & 2.7 & 0.1 & 0.1 & 0.1\\
Weight assigned & 0.01 & 0.1 & 0.7 & 0.7 & 0.7\\
\bottomrule
\end{tabular*}
\end{table}

\subsection{Training settings}  The model is trained using Adam optimizer with a learning rate starting with a value of 1e-4, which is gradually reduced by a factor of 0.6 following a multi-step learning rate scheduler. For the xBD dataset, the model takes in 1,024 $\times$ 1,024 images as input and is trained using a batch size of 8 for 50 epochs. For the LEVIR dataset, the model takes in 256 $\times$ 256 cropped images and is trained using a batch size of 16 for 200 epochs. {Geometric and photometric data augmentations are heavily used to make the model robust to variations and noise in the dataset. Further, small random shifts and rotation of images are used to handle any shift in pixels among the imagery pair}. A 10\% of the data is held for validation and the model is trained on the remaining set. The training has been done using p3.8xlarge instances of Nvidia GPUs on an AWS cloud environment and took nearly 6 hours for 200 epochs on xBD dataset.

\subsection{Domain adaptation}\label{3.1.3} This paper demonstrates that the model is capable of handling other datasets using a simple transfer learning approach. This also shows a path on how to make use of small datasets for other disasters including the one we have provided (Ida-BD, introduced in Section \ref{sec42}). The diverse nature of satellite imagery captured from different locations, terrains, and weather conditions makes machine-learning algorithms difficult to generalize, which means the model trained on the xBD dataset does not generalize well for other datasets. Additionally, training a model on small datasets from scratch is difficult for the highly complex tasks such as damage classification. Several domain adaptation algorithms have been proposed \citep{domadapt_survey} to enable models trained on images from one dataset (source) to work on images from another dataset (target). Here Ida-BD dataset is used as target domain and since it has a domain shift from the xBD dataset, this work applies the discrepancy-based method to fine-tune the network with a small labeled target data. Since the percentage of destroyed category is relatively lower than the rest of the classes (Table \ref{tab_xbd_class_dist}), the destroyed class is merged with major damage class. This is handled by replacing the output layer head by 4 neurons head instead of 5 neurons and thus framing the problem as four-class classification (i.e., background, no damage, minor damage, major damage). As a result, Ida-BD dataset also poses the challenge of concept shift (5-class to 4-class mapping) along with the domain shift (different terrain). Multiple settings are experimented to effectively train a model on Ida-BD dataset which include using the pretrained model directly from xBD dataset on the target domain (works with unlabelled data), fine-tuning the pre-trained model on the target dataset (works only after some labelling) and training a model from scratch (works only after some labelling). The performance is compared in the Section \ref{sec5.2.2}. For fine-tuning experiments, the pre-trained model is trained on Ida-BD dataset for 10 epochs using a fixed learning rate of 1e-6.

\section{Datasets}\label{sec4} The xBD dataset \cite{xBDdataset} was used for model training and testing for disaster damage detection and classification tasks. Later, a new dataset of higher resolution (but smaller size) is introduced in this paper from Hurricane Ida for damage assessment and explore the domain adaptation from xBD to this data. Additionally, this work also uses LEVIR-CD dataset to evaluate the model for the change detection tasks. In all these datasets, the images from pre-event and post-event are coordinate-registered by the imagery providers.

\subsection{xBD}\label{sec4.1} This work uses the xBD dataset \citep{xBDdataset} for disaster damage detection and classification tasks. The xBD dataset is a large-scale building damage dataset that publicly offers high-resolution (0.8 m/pixel) satellite imagery with building segmentation and damage level labels. The xBD dataset provides images paired from pre- and post-disaster images collected from 19 disaster events such as floods and earthquakes with an image size of 1,024 $\times$ 1,024 pixels. The dataset uses polygons to represent building instances and provides four damage categories --- no damage, minor damage, major damage and destroyed --- for each building. As described in Table \ref{class_description}, the minor damage pixels represent visible roof cracks or partially burnt structures while the major damage represents partial wall, roof collapse or structure surrounded by water. The destroyed label mean that the building structure has completely collapsed, scroched or is no longer present. The work makes use of the tier 1 and tier 3 data partitions where the labels are available. Using stratified sampling based on the damage class distribution, 10\% of this data is kept aside for validation, and another 10\% for final testing. The rest of the data is used for training the model.
\begin{table}[!htbp]
\centering
\caption{Class-wise representation of damage \citep{xBDdataset}.}\label{class_description}%
\begin{tabular*}{350pt}{@{\extracolsep\fill}ll}
\toprule
\textbf{Class name} & \textbf{Description} \\
\midrule
No damage & Undisturbed. No sign of water, structural or \\ & burn damage.\\
Minor damage & Building partially burnt, water surrounding\\
& structure, roof elements missing or \\ & visible cracks.\\
Major damage & Partial wall or roof collapse, water surrounded. \\
Destroyed & Completely collapsed, completely covered \\ & with water/mud or no longer present.\\
\bottomrule
\end{tabular*}
\end{table}
\subsection{Ida-BD}\label{sec42}
\begin{figure*}
\centerline{\includegraphics[width=0.9\linewidth]{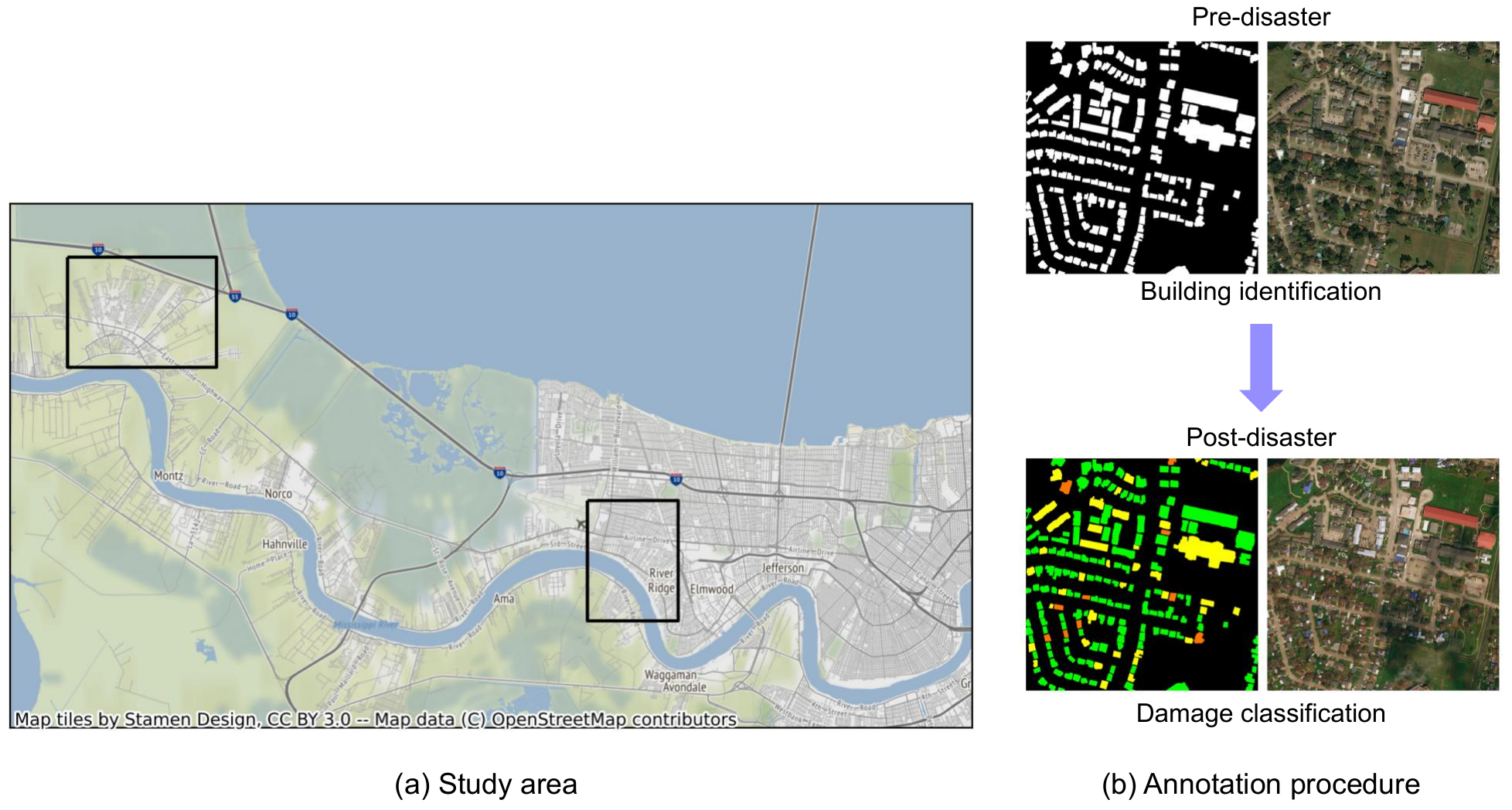}}
\caption{(a) Location of Ida-BD near New Orleans, Louisiana\label{studyarea}, and (b) example of the annotation of buildings and damage levels.}
\end{figure*}

\begin{figure*}
\centerline{\includegraphics[width=\linewidth]{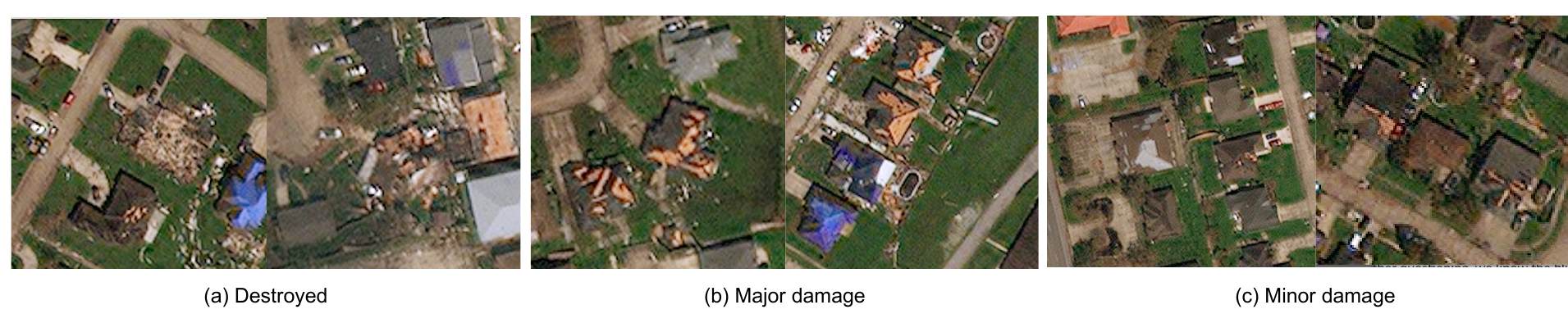}} 
\caption{Examples of buildings identified as destroyed, major damage, and minor damage in Ida-BD dataset.}\label{damage_example}
\end{figure*}

A new dataset, Ida-BD, obtained from the WorldView-2 (WV2) satellite with very-high-resolution images taken in November 2020 and July 2021 for pre-disaster and September 2021 for post-disaster is offered along with this paper. The satellite imagery was collected close to New Orleans, Louisiana, one of the most heavily impacted areas during Hurricane Ida in late August 2021 (Figure \ref{studyarea}a). The WV2 satellite provides panchromatic images with spectral resolution of 450--800 nm and spatial resolution of 46 cm. The panchromatic images in this dataset were first orthorectified by Apollo Mapping to 0.5 m/pixel. The work then created 87 image pairs (pairs of pre- and post-disaster images at the same location) with a size of 1,024 $\times$ 1,024 pixels. The resolution of these images is finer than the ones in the xBD dataset. Similar to xBD, polygons were used to represent building segments and provided four damage categories. All annotations were done by the in-house team with quality control procedures using Labelbox \citep{labelbox_labelbox_2022}. The annotation procedure is shown in Figure \ref{studyarea}b. Building polygons were first annotated in pre-disaster images to avoid incorrect building boundaries due to damage. Then, by overlapping the annotation of building boundaries, building damage was classified for each building based on post-disaster images. All annotations, including building boundaries and damage levels, were reviewed several times by experts on the team. The damage levels were labelled as per the criterion introduced by the xBD dataset and are specified as --- no damage, minor damage, major damage and destroyed --- for each building. Figure \ref{damage_example} shows the examples of building identified as destroyed, major damage, and minor damage in Ida-BD dataset.

The comparison of damage class distribution in xBD and Ida-BD datasets is shown in Table \ref{tab_xbd_class_dist}. Also, the statistical summary of Ida-BD is presented in Table \ref{ida_bd_summary}. Ida-BD shows more damaged buildings at all levels in terms of pixel counts compared with xBD except for the class of destroyed buildings. Due to the small size of the dataset, it is hard to use it directly for training a model from scratch; therefore, this work leveraged domain adaptation using a pre-trained model on the xBD dataset for training and evaluation on this dataset. Since the xBD dataset includes various disaster types, the class distribution of xBD dataset is quite different from Ida-BD dataset. To address this shift in target distribution, updated the class weights are updated in the loss function according to the new distribution. Also, aggressive data augmentation is used for scaling and normalization to model transfer knowledge from the dataset with different resolution and RGB distribution. We have made this dataset publicly available at DesignSafe-CI \citep{lee_ida-bd_2022}.

\begin{table}[!htbp]
\centering
\caption{Class-wise pixel percentage comparison of xBD dataset with Ida-BD dataset. The notation used: 0--Background, 1--No damage, 2--Minor damage, 3--Major damage, 4--Destroyed.\label{tab_xbd_class_dist}}%
\begin{tabular*}{400pt}{@{\extracolsep\fill}lccccc@{\extracolsep\fill}}%
\toprule
\textbf{Class Id} & {\textbf{0}} & {\textbf{1}} & {\textbf{2}} & {\textbf{3}} & {\textbf{4}} \\
\midrule
xBD & 96.1  & 2.7 & 0.1 & 0.1 & 0.1\\
Ida-BD & 81.7 & 11.9 & 4.6 & 1.6 & 0.05\\
\bottomrule
\end{tabular*}
\end{table}

\begin{center}
\begin{table*}[!htbp]
\centering
\caption{Statistical summary of Ida-BD dataset. \label{ida_bd_summary}}%
\begin{tabular*}{300pt}{@{\extracolsep\fill}lccc@{\extracolsep\fill}}%
\toprule
\textbf{Damage class} & \textbf{Building polygons} & \textbf{Pixel count} & \textbf{Pixel \%} \\
\midrule
No damage & 13,667 &  10,879,251 & 11.9 \\
Minor damage & 3,247 & 4,214,728 & 4.6 \\
Major damage & 1,120 & 1,524,746 & 1.6 \\
Destroyed & 49 & 52,884 & 0.05 \\
\bottomrule
\end{tabular*}
\end{table*}
\end{center}

\subsection{LEVIR-CD} 
LEVIR-CD consists of bi-temporal satellite images from Texas capturing significant land-use changes such as new construction or building decline. This dataset has 637 pairs of very-high-resolution (VHR, 0.5 m/pixel) satellite imagery with periods of 5 to 14 years and a size of 1,024 $\times$ 1,024 pixels. There are nearly 31K building instances within this dataset, including various building types such as warehouses, garages, and apartment complexes. This dataset provides binary masks as labels with values 1 for changed and 0 for unchanged. Similar to the split for the xBD dataset, this work keeps aside 10\% of the data for validation and another 10\% for testing while using the rest for model training.
\section{Evaluation}\label{sec5}
In this section, the paper presents the performance of the model on the datasets listed in Section \ref{sec4}. The metrics used for the evaluation is shared, and the qualitative and quantitative results are presented. The impact of components of our model and loss functions are examined through ablation studies.
\subsection{Metrics}
The results are evaluated on xBD dataset using the combined F1-score metrics from the XView2 Challenge 2019 \citep{xBDdataset}. The metrics include a weighted average of the F1-scores of building segmentation results and the harmonic mean of the F1-scores of building damage classification results. The higher weight is assigned to the classification score since it is a multi-class classification problem with highly skewed labels and hence more challenging to achieve.
\begin{equation}\label{eq23}
Score = 0.3 * F1_{Loc} + 0.7 * F1_{Class}
\end{equation}
\begin{align*}
F1_{Loc} & =  \frac{2 \mid X \cup Y \mid}{\mid X\mid \cap \mid Y\mid } \\
F1_{Class} & = \frac{1}{\frac{1}{F1_i}+\dots+\frac{1}{F1_n}}
\end{align*}
Here, $F1_{Loc}$ is the F1-Score for building segmentation. X and Y denote background and buildings interchangeably.
$F1_i$ denotes class-specific F1 scores for each damage level. Since this metric heavily penalizes overfitting on the over-represented classes and given that xBD dataset is heavily skewed towards no-damage class, it is a challenging metric to use for evaluation.

\subsection{Quantitative and qualitative results}
Here the results are discussed for the three tasks: damage detection and classification, change detection, and domain adaptation tasks. This work evaluates the results for these tasks on xBD dataset (damage classification), LEVIR dataset (change detection), and Ida-BD (domain adaptation) dataset.

\subsubsection{Results for damage classification}
\begin{figure*}
\centerline{\includegraphics[scale=0.7]{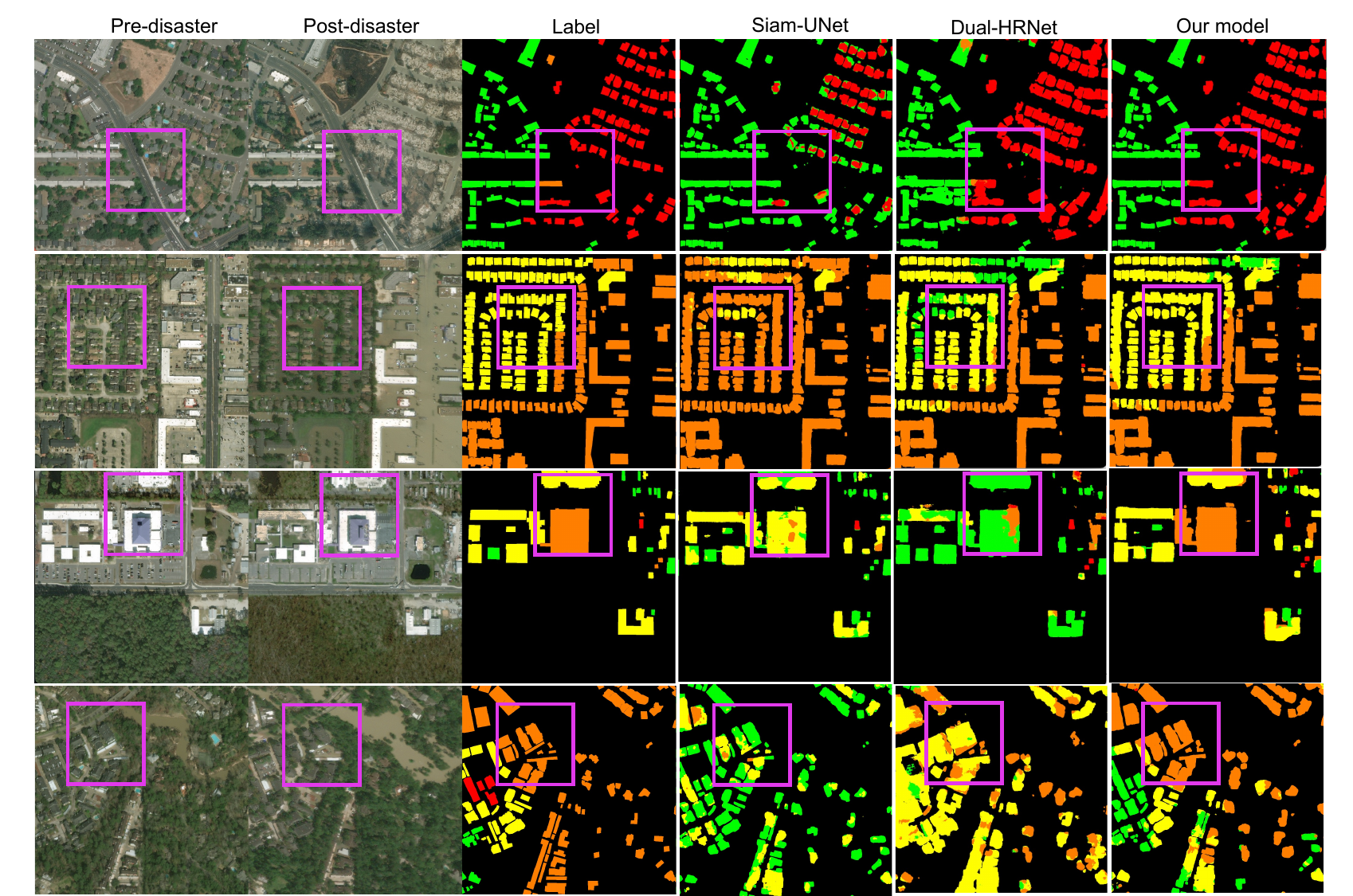}}
\caption{Qualitative results for damage classification (evaluation on xBD dataset). As highlighted in the boundary boxes, our model assigns more accurate damage level to almost all pixels within a building boundary as compared to Siamese UNet and Dual-HRNet\label{fig_xbd_results}}
\end{figure*}

\begin{figure*}
\centerline{\includegraphics[scale=0.85]{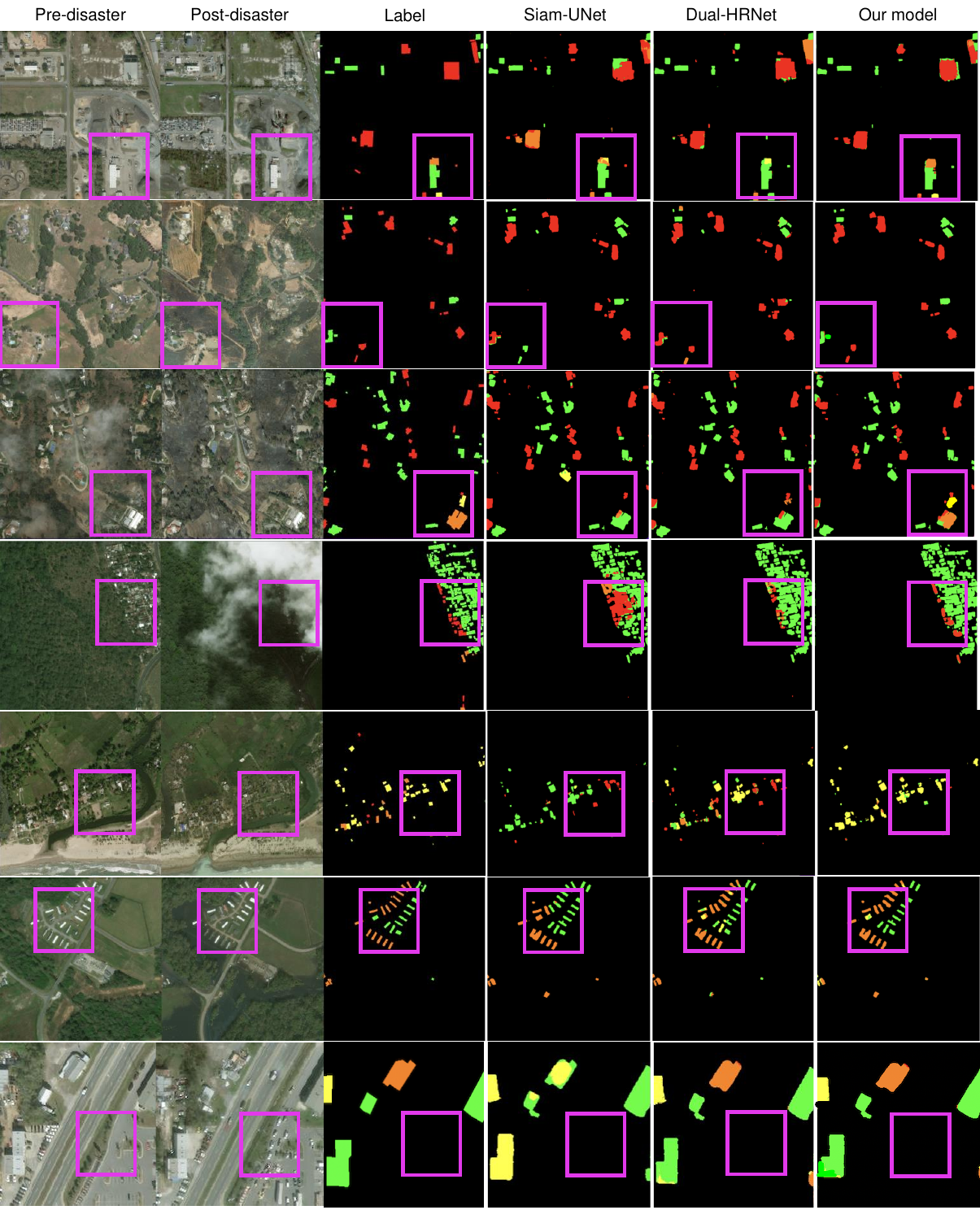}}
\caption{Additional figures for qualitative results for damage classification (evaluation on xBD dataset), containing images from hard terrains, images with poor quality and rural regions. As observed in the figure, the model performs well on these images as compared to other baselines. The last row contains zoomed in results from imagery with moving objects like vehicles. The models effectively ignore such objects while identifying the building structures.\label{fig_xbd_results_poor}}
\end{figure*}

The performance of the proposed model for the damage detection and classification tasks on the xBD dataset is evaluated by comparing the F1-score and IoU metrics. The results are compared with multiple benchmarks and recent works.
\begin{enumerate}
    \item Two-stage CNN based models: The Siamese UNet \citep{unet} is a fully convolutional and siamese based network and generates segmentation mask and classification mask in two stages. RescueNet \citep{rescuenet} uses a dilated ResNet and atrous spatial pyramid pooling framework to generate multi-scale features and then uses independent heads for segmentation and classification tasks.
    \item Fusion based models: The Dual-HRNet \citep{dualhrnet} uses fusion blocks to exchange the features of pre-disaster and post-disaster images. The BDANet \citep{bdanet} uses cross-directional attention to exchange the features of pre-disaster and post-disaster images.
    \item Transformer based model: BiT \citep{chen2021bit} uses a single transformer encoder to learn the temporal context between the pre-disaster and post-disaster, project the features back to the pixel-space and then generate the classification map using upsampling.
\end{enumerate}
As presented in Table \ref{tab_xbd_results}, the proposed model outperforms the existing methods for the damage classification task. Firstly, since the Siamese UNet \citep{siam-att-unet} is based on comparison of pre- and post- features at the very end of the model layers, it does not perform well for the classification task (gain of 0.076 is observed with the proposed model). Since the RescueNet \citep{rescuenet} model also concatenates the features from pre- and post- layers at the very end of the model, it provides comparatively low final score (gain of 0.053 is observed with the proposed model) The Dual-HRNet \citep{dualhrnet}, which is one of the top-performing architectures in xview2-challenge, also provides low score for F1 metrics which is reasonable because it does not account for multi-resolution features during comparison or feature construction. (gain of 0.05 is observed with the proposed model). The BDANet \citep{bdanet} incorporates attention based exchange of information from pre- and post- and uses multi-resolution input features and performs well for major and destroyed classes; however, the proposed model provides a gain of 0.013 in the overall score by using richer features extracted via transformers. The BiT uses a single transformer model and uses one-step upsampling to generate the final mask which makes it difficult to get a good IOU score as well as f1-score for high resolution images. The qualitative results are displayed in Figure \ref{fig_xbd_results}, where the proposed model assigns the same damage level to almost all pixels within a building boundary, in contrast to other models. Figure \ref{fig_xbd_results_poor} presents additional results from rural regions and images with poor quality. Please note that the damage classification problem is significantly difficult for the images with very good quality. Further addition of noise from clouds and dust makes it more difficult to get accurate results. Though there are some performance gaps on the images with clouds (rows 3 and 4) in the precision of the building boundaries, the proposed model performs decent comparatively to the other models. Few challenges posed in the damage classification on satellite imagery that have not been addressed in this work include the damages which are not visible clearly in the images. For example, the buildings that are covered under trees before or after the disaster or the damaged regions which are fully covered under the clouds. Additionally, the last row in \ref{fig_xbd_results_poor} presents zoomed in results for an imagery with moving objects like cars. All the models effectively ignore such objects and focus more on identifying building structures. This is because the xBD dataset has been labeled only for the building polygons and hence the models learn to distinguish building structures from rest of the moving objects.

\begin{center}
\begin{table*}[t]%
\centering
\caption{Average quantitative results for damage classification on xBD dataset. Our model outperforms the state-of-the-art methods in the overall IoU and F1-score metrics as well as class-specific F1-scores for minor damage class and no-damage class. Although BDANet performs better for major damage and destroyed classes, our model has higher overall scores. \label{tab_xbd_results}}%
\begin{tabular*}{470pt}{@{\extracolsep\fill}lccccccc@{\extracolsep\fill}}
\toprule
& & & & \multicolumn{4}{@{}c@{}}{{Class F1-scores}} \\\cmidrule{5-8}
  \textbf{Model} & \textbf{Score}  & \textbf{IoU} & \textbf{F1-Score} & {\textbf{No Damage}}  & \textbf{Minor Damage} & \textbf{Major Damage} & \textbf{Destroyed}\\
\midrule
Siam-UNet & 0.743 & 0.824  & 0.709  & 0.955 &  0.576  & 0.744 & 0.662 \\
RescueNet & 0.766 & 0.840  & 0.735  & 0.883 & 0.563 & 0.771 & 0.808\\
Dual-HRNet & 0.769 & 0.834 & 0.741 & 0.898 & 0.590 & 0.737 & 0.809\\
BDANet & 0.806  & 0.864  & 0.782  & 0.925  & 0.616  & \textbf{0.788} & \textbf{0.876}\\
BiT & 0.764 & 0.816 & 0.742 & 0.971 & 0.631 & 0.723 & 0.719  \\
Ours & \textbf{0.819} &  \textbf{0.872} & \textbf{0.796} & \textbf{0.978} & \textbf{0.711} & 0.765 & 0.772 \\
\midrule
ChangeFormer {$^{\rm a}$} & 0.458 & 0.678 & 0.364 & 0.778 & 0.389 & 0.493 & 0.196 \\
Ours {$^{\rm a}$} & 0.495 & 0.734 & 0.393 & 0.725 & 0.412 & 0.487 & 0.232\\
\bottomrule
\end{tabular*}
{\raggedright \item[$^{\rm a}$] Here the models are trained with batch-size of 1 due to GPU compute constraints for ChangeFormer. The rest of the models are trained with a batch size of 8. Please refer to Table \ref{performance_t} for the computational performance comparison. \par}
\end{table*}
\end{center}

\subsubsection{Results for change detection}
The performance of change detection is evaluated on LEVIR dataset and compared against the following methods:
\begin{enumerate}
    \item CNN based model: The Siamese UNet \citep{unet} is a fully convolutional and siamese based network and is used as a simple baseline for the comparison.
    \item Transformer based models: BiT \citep{chen2021bit} uses a single transformer encoder to learn the temporal context between the pre-disaster and post-disaster, project the features back to the pixel-space and then generate the classification map using upsampling. Changeformer \citep{changeformer} uses hierarchically structured transformer encoder modules along with multi-layer perceptron (MLP) decoders.
\end{enumerate}

\begin{figure*}
\centerline{\includegraphics[scale=0.55]{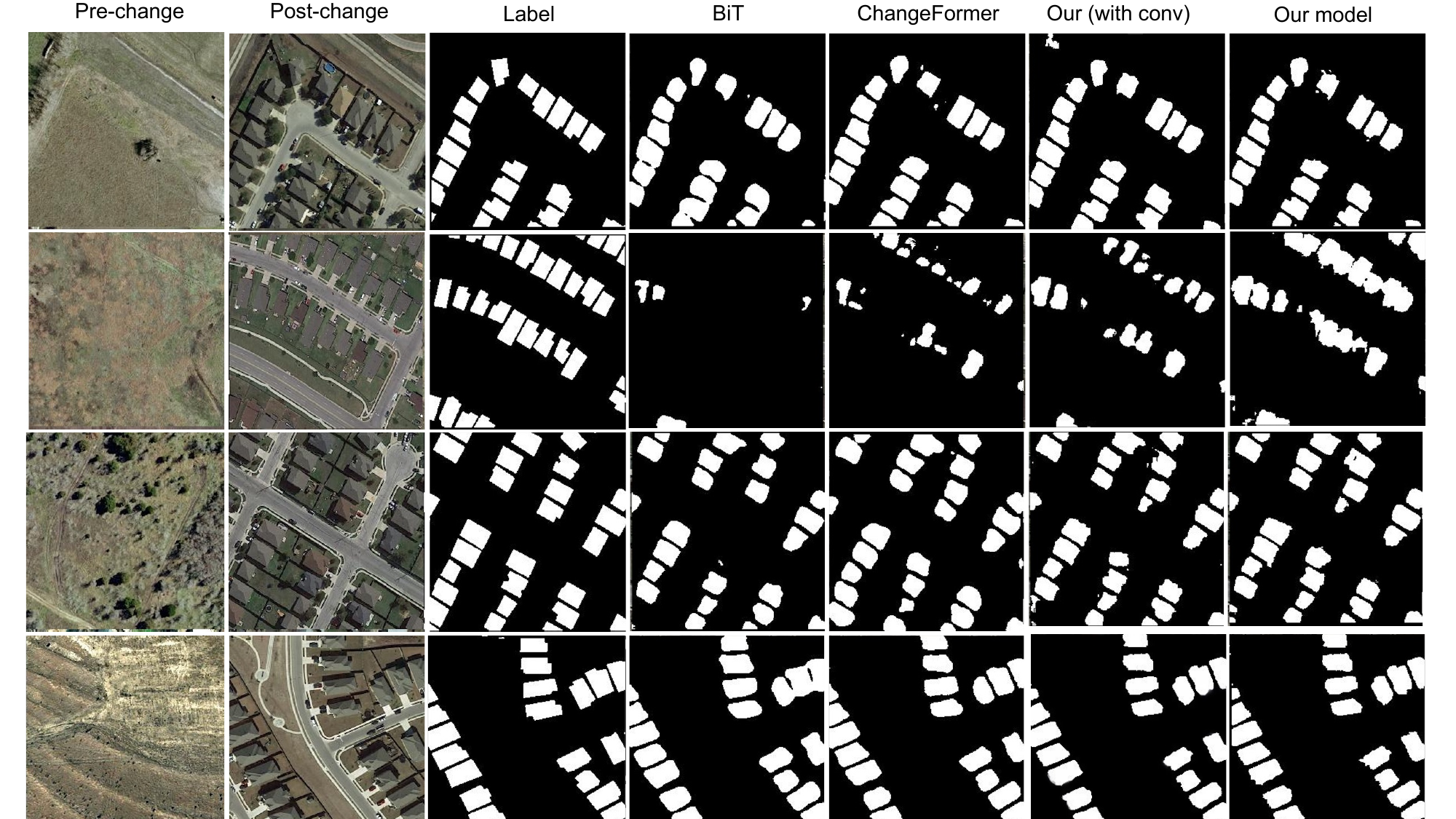}}
\caption{Qualitative results for change detection (evaluation on LEVIR-CD dataset.)\label{fig_levir_results}}
\end{figure*}

The proposed model performs significantly better in both overall F1-score and IoU metrics, as demonstrated in Table \ref{tab_levir_results}. Since the BiT model uses a single transformer encoder and decoder pair and then generate the output mask by up-sampling the decoder features, it is difficult to get a high resolution pixel-level classification mask. On the other hand, Changeformer uses hierarchically structured transformer encoder modules but the features are not projected back into the spatial domain using the transformer decoders. Instead, the proposed network explicitly takes the difference in the transformer encoded feature domain and then maps back the features into the spatial domain to re-construct the final output. This leads to a better performing framework for change detection. In addition, adding a convolution layer after every merge of outputs from the transformer decoder and lower layer provides smoother results and fewer artifacts. The accuracy achieved under this setting is slightly lower than our final model, though higher than other recent works. The qualitative results are displayed in Figure \ref{fig_levir_results}, where it can be seen that the model produces finer building boundaries and captures a few additional buildings that had been missed by other models.

\begin{center}
\begin{table*}[t]%
\centering
\caption{Average quantitative results for change detection on LEVIR-CD dataset. Our model performs significantly better than the state-of-the-art methods in overall F1-score and IoU as well class-specific F1-scores.\label{tab_levir_results}}%
\begin{tabular*}{400pt}{@{\extracolsep\fill}lcccc@{\extracolsep\fill}}
\toprule
& & & \multicolumn{2}{@{}c@{}}{{Class F1-scores}} \\\cmidrule{4-5}
  \textbf{Model}  & \textbf{IoU} & \textbf{F1-Score} & \textbf{0 class} & \textbf{1 class}\\
\midrule
Siam-UNet & 0.813  & 0.859  & 0.987 & 0.788   \\
BiT	&	0.829	& 0.899	&	0.990 &	0.807   \\
Changeformer  &	0.828	&	0.898	&	0.990	&	0.806   \\
Ours + with conv &	0.833	& 0.901	&	0.991 &	0.812\\
Ours  & \textbf{0.842}   & \textbf{0.908}  & \textbf{0.991}  & \textbf{0.825}  \\
\bottomrule
\end{tabular*}
\end{table*}
\end{center}

\subsubsection{Results for domain adaptation}\label{sec5.2.2}
\begin{figure*}
\centerline{\includegraphics[scale=0.72]{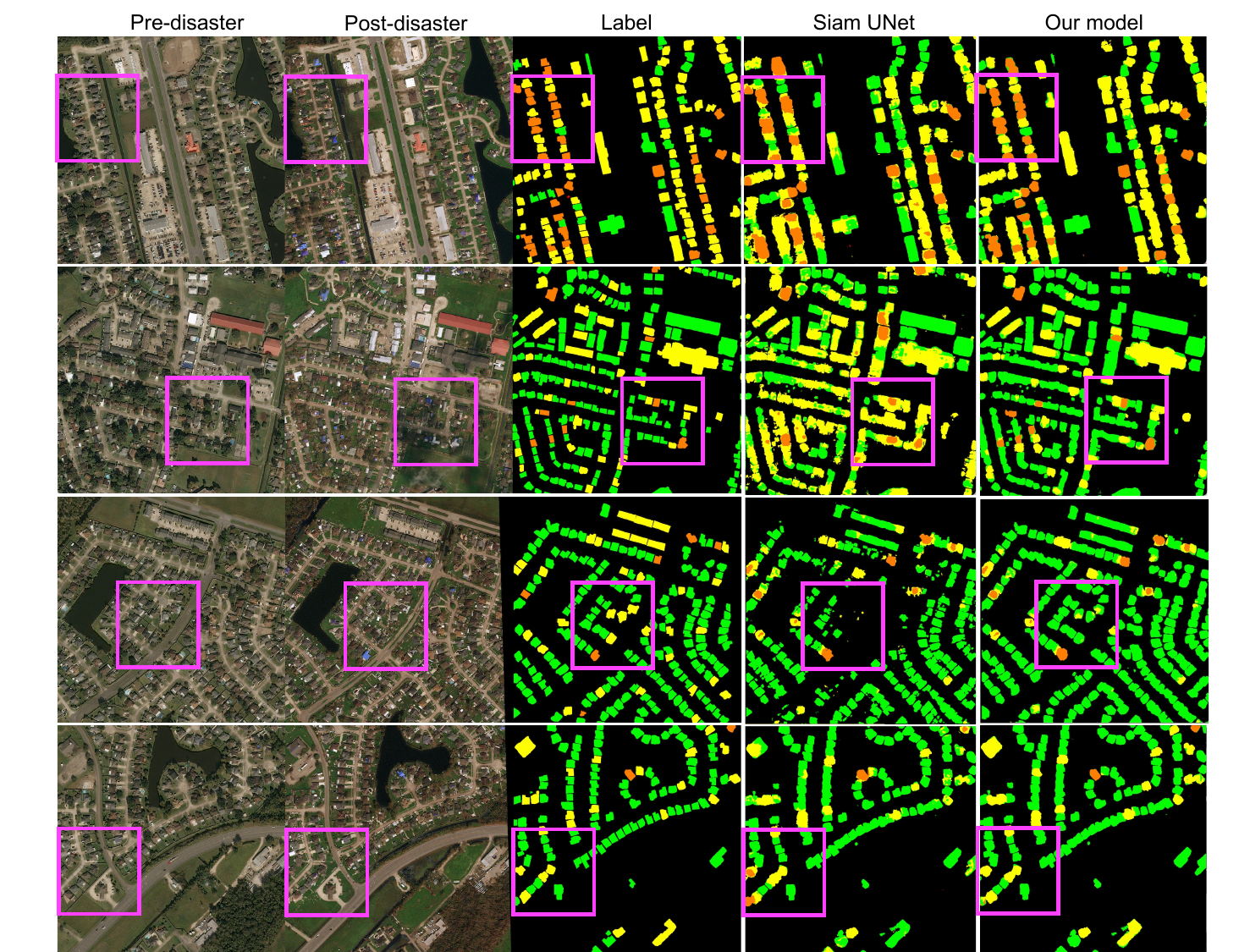}}
\caption{Qualitative results for domain adaptation (evaluation on Ida-BD dataset). Our model assigned significantly more accurate damage classes than Siamese UNet. Please compare the highlighted boxes as a reference.\label{fig_DA_results}}
\end{figure*}

The purpose of this section is to demonstrate a simple approach to using the model on new and small datasets like Ida-BD. Please note that the model is not necessarily better than the rest of the models for this task and the authors only present one of the methods for transferring learning on new datasets. To evaluate the model adaptation performance on the Ida-BD dataset, this work experiments with three training settings for our model and the Siamese UNet as a baseline model. For both the models, first, this work trained the model on Ida-BD data from scratch and tested it. In the second setting, this work initializes the models with weights from the pre-trained model on xBD data and compared their performance. Finally, this work evaluated the pre-trained model directly without any fine-tuning on the Ida-BD dataset. As shown in Table \ref{tab_domainadapt_results} for both the models, the best performance was obtained by initializing the model with weights from the network trained on xBD data and then fine-tuning on the Ida-BD dataset. This shows that to use models for small datasets like Ida-BD, pre-training on the large-scale xBD dataset provides a good prior, and hence models are able to better generalize to new datasets. The qualitative results are displayed in Figure \ref{fig_DA_results}.

\begin{center}
\begin{table*}[ht!]%
\centering
\caption{Domain adaptation on Ida-BD dataset. This work experimented using three settings: (1) This work trained the model on Ida-BD data (source and target) from scratch. (2) This work initialized the model with weights from the pre-trained model on xBD data (source) and fine-tune on Ida-BD data (target). (3) This work tested the model trained on xBD data (source) on Ida-BD (target) without target knowledge. The second setting performs the best for both the models while our model performs better than Siam-UNet in each of the settings in terms of F1-score and IoU. \label{tab_domainadapt_results}}%
\begin{tabular*}{470pt}{@{\extracolsep\fill}lccccccc@{\extracolsep\fill}}
\toprule
& & & & \multicolumn{4}{@{}c@{}}{{Class F1-scores}} \\\cmidrule{6-8}
  \textbf{Model} & \textbf{Pre-training}  & \textbf{Training}  & \textbf{IoU} & \textbf{F1-Score} & {\textbf{No Damage}}  & \textbf{Minor Damage} & \textbf{Major Damage}\\
\midrule
Siam-UNet & - & Ida & 0.697	& 0.472 &	0.906 &	0.313 &	0.483 \\
 & xBD & Ida & 0.748 & 0.507 &	0.846 &	0.322 &	0.609 \\
 & xBD & - & 0.584 & 0.307 & 0.916 & 0.208 & 0.251 \\
\midrule
Ours & - & Ida & 0.778 & 0.541 & \textbf{0.916} & 0.384 & 0.538 \\
 & xBD & Ida & \textbf{0.805} & \textbf{0.585} & 0.910 & 0.439 & \textbf{0.577} \\
 & xBD & - & 0.674 & 0.023 & 0.881 & \textbf{0.617} & 0.008 \\
\bottomrule
\end{tabular*}
\end{table*}
\end{center}

\subsubsection{Limitations}
This work has used with a simple domain adaptation approach for the Ida-BD dataset. Further experiments can be done using the domain adaptation methods like knowledge distillation \citep{knowledge_distill}, few-shot learning \citep{few-shot-learning} and adversarial-based domain adaptation using generative models (GANs) \citep{GAN_domain_adapt}.

\subsection{Ablation studies}
In this sub-section, the paper discusses few experiments on the performance and impact of distortions and different loss functions on our model.

\subsubsection{Impact of distortion}
The pre-event and post-event images from the xBD, LEVIR-CD and Ida-BD datasets are coordinate-registered by the imagery providers and hence they are perfectly aligned. However, the impact of slight translation or angular deviation is discussed by training on images with random translational distortion and angular deviation. As presented in Table \ref{distortion}, training the model on images randomly translated up to 1m gives a small decrease of 0.008 in F1-score and 0.003 IOU metrics. Similarly, angular deviation uptil 0.1 degrees gives a very small decrement of 0.004 in F1-score and 0.004 IOU metrics. This could be explained using the property of transformers that it learns global features and not limited to local context. However, further distortions upto 2 meters or angular deviation of 1 degree leads to F1 score decrement of 0.03 and 0.05 respectively.

\begin{table}
\centering
\caption{Impact of translational distortion (m) and angular deviation (degrees) on partial xBD dataset \label{distortion}}%
\begin{tabular*}{320pt}{@{\extracolsep\fill}cccc@{\extracolsep\fill}}%
\toprule
\textbf{Translation} & \textbf{Angular deviation} & \textbf{IOU} & \textbf{F1-Score} \\
\midrule
- & - & 0.872 & 0.839 \\
1 & - & 0.869 & 0.831 \\
2 & - & 0.865 & 0.803 \\
- & 0.1 & 0.868 & 0.835 \\
- & 1 & 0.861 & 0.786 \\
\bottomrule
\end{tabular*}
\end{table} 

\subsubsection{Computational analysis}
The computational cost and inference speed analysis of the network is compared with the recent models. As seen in Table \ref{performance_t}, the transformer based models BiT and \dahitra{} have lower number of parameters than CNN models. The inference speed of Siam-UNet is the lowest since it is fully convolutional and can be parallelized more efficiently. However, the transformer based models are faster than fusion based models including BDANet and Dual-HRNet.

\begin{table}
\centering
\caption{Computational analysis of the network\label{performance_t}}%
\begin{tabular*}{320pt}{@{\extracolsep\fill}lccc@{\extracolsep\fill}}%
\toprule
\textbf{Model} & \textbf{Params (M)} & \textbf{Inference (im/s)} & \textbf{FLOPs (G)} \\
\midrule
Siam-UNet & 25.7 & 0.620 & 26.37\\
BDANet & 34.5 & 0.681 & 82.22\\
{ChangeFormer} & {41.1} & {2.832} & {209.41}\\
BiT & 12.4 & 0.636 & 11.52\\
Ours & 13.2 & 0.625 & 11.95\\
\bottomrule
\end{tabular*}
\end{table} 

\subsubsection{Dataset choice for domain adaptation}\label{idabd_xbd_val}
As discussed in the introduction, adapting the model to new imagery is challenging and there are two different alternatives for choosing a dataset to study domain adaptation. One method is to use imagery from one of the disaster events in xBD dataset as test data while training on the remaining imagery. Another method is to use datasets like Ida-BD as a test set, which is from a different imagery source and different resolution, and train on xBD dataset. In this experiment, we first assumed that no labeled data is available for the `new' disaster event and compared the models with no fine-tuning on the `new' disaster class. Next, we have compared the performance by assuming that 30\% of imagery for the `new' disaster event is labeled and hence the model is fine-tuned using 30\% of imagery from the `new' disaster class. {As seen in Table \ref{idabd_xbd_val_t}, in the absence of the training data for fine-tuning, the model does not work well in both cases but a small amount of fine-tuning improves the performance significantly for both the datasets. The test performance for disaster damage classification is lower for Ida-BD imagery than ‘Hurricane Matthew’ imagery for both experiments which could be because the  ‘Hurricane Matthew’ imagery is more similar to the training dataset. This is also indicative that adapting the model for damage classification to new imagery from a different imagery source and different resolution (here Ida-BD) is more challenging. On the other hand, the building segmentation results are better for the Ida-BD dataset for both experiments which could be due to the higher resolution of Ida-BD imagery.}

\begin{table}
\centering
\caption{Performance of \dahitra{} for different choices of the dataset for domain adaptation study. Here xBD-T stands for xBD dataset except for Hurricane Matthew imagery and xBD-V stands for xBD dataset except for Hurricane Matthew imagery, along with 30\% imagery of test class \label{idabd_xbd_val_t}}%
\begin{tabular*}{320pt}{@{\extracolsep\fill}lccc@{\extracolsep\fill}}%
\toprule
\textbf{Test} & \textbf{Train} & \textbf{IoU} & \textbf{F1-Score} \\
\midrule 
Hurr. Matthew (xBD) & xBD-T & 0.553 & 0.191\\
Ida-BD & xBD-T & 0.628 &  0.026\\
Hurr. Matthew (xBD) & xBD-V & 0.720 & 0.724\\
Ida-BD & xBD-V & 0.773 & 0.578\\
\bottomrule
\end{tabular*}
\end{table}

\subsubsection{Loss functions}
This work experimented with different combinations of loss functions (Table \ref{tab_loss_fn}). The localization-aware loss \citep{rescuenet} gave better F1-scores for damage classes where pixels are accounted for in the cross-entropy loss calculation only in the presence of buildings. The ordinal loss is also experimented in the form of mean square error between the ground truth and predicted output for buildings only by converting the values from a scale of 1 to 4 to the scale of 0 to 1. This loss did not work well for our task, which could be because of the highly imbalanced and sparse class distribution. The weighted sum of focal and dice loss provided the best performance in class-wise F1-scores as well as building boundary precision, since the former explicitly addresses the class imbalance and the latter ensures crisp building boundaries.

\begin{table}
\centering
\caption{Impact of loss functions (xBD data)\label{tab_loss_fn}}%
\begin{tabular*}{320pt}{@{\extracolsep\fill}lcc@{\extracolsep\fill}}%
\toprule
\textbf{Loss function} & \textbf{IoU} & \textbf{F1-Score} \\
\midrule
Focal + Dice & \textbf{0.816} & 0.796 \\
Focal + Dice + Ordinal & 0.807 & 0.791 \\
Buildings only Cross-entropy & 0.798 & \textbf{0.803}\\
\bottomrule
\end{tabular*}
\end{table} 

\subsubsection{Transformer layers}
This work experimented with adding different numbers of transformer blocks between encoder and decoder; three iterations gave us the best results. It can be observed that adding transformer blocks at all levels, except the highest resolution layer, provided an incremental improvement in accuracy. An addition transformer block in final layer does not perform well, which could be because the input dimension is very high and it becomes hard for transformers to learn in this space, as discussed in \citet{Wu_2021_ICCV}. Hence this work used only the convolutional layer in the highest resolution layer. On the other hand, if the transformer is not used in any of the layers, the performance highly degrades, indicating the importance of transformer-encoded features. The test result values from the LEVIR-CD dataset are shared in the Table \ref{tab_transformer_layer}.  Also, using a combination of transformer encoder and decoder along with the difference module, instead of just transformer encoder followed by difference module, provided a bump of accuracy by 0.8\% for xBD dataset.

\begin{table}
\centering
\caption{Effect of the number of transformer layers used (LEVIR data)\label{tab_transformer_layer}}%
\begin{tabular*}{320pt}{@{\extracolsep\fill}ccc@{\extracolsep\fill}}%
\toprule
\textbf{Number} & \textbf{IoU} & \textbf{F1-Score} \\
\midrule
0 & 0.818 & 0.843 \\
1 & 0.837 & 0.897 \\
2 & \textbf{0.843} & 0.903 \\
3 & 0.842 & \textbf{0.908}\\
4 & 0.838 & 0.901\\
\bottomrule
\end{tabular*}
\end{table}
\section{Concluding Remarks}\label{sec6}
This paper presents a hierarchical UNet-based architecture that uses transformer-based differences to perform well on both the damage classification and building segmentation tasks. This is the first method that explicitly relies on the difference between transformer-encoded features from pre-disaster and post-disaster images and hierarchically builds the output from multi-resolution features by using UNet as a backbone. {The proposed method yields state-of-the-art results on xBD dataset and LEVIR-CD dataset for damage classification and change detection tasks respectively.  Additionally, this work introduces a new dataset, Ida-BD, for evaluating the model's adaptation performance and providing a baseline for the damage assessment tasks.} It also demonstrates the usefulness of the new dataset and the efficacy of the proposed network using transfer learning. The domain adaptation results indicate that the model can be adapted to a new event with slight fine-tuning, which is critical for the application of the model for future events. In addition to application to building damage assessment, the proposed network architecture with hierarchical transformers can be used in other civil infrastructure and urban systems applications; for example, this model architecture is expected to perform well on problems such as road damage classification, structure change detection, and urban land cover change classification problems.

In future work, the authors plan to improve the building boundaries using either GAN loss or exponential boundary loss, which could help to split different building instances. The work can be further extended by using dynamic algorithms like neural dynamic classification \citep{neuraladelhi} and dynamic ensemble learning \citep{dynamicensemble} and can be  optimized for faster learning using finite element machines \citep{fema}.  

\section{Acknowledgments}
The authors would like to acknowledge funding support from the Texas A\&M X-Grant Presidential Excellence Fund. 

\section{Code Availability}
The code of the proposed model in this study is available at \href{https://github.com/nka77/DAHiTra}{https://github.com/nka77/DAHiTra}

\section{Data Availability}
xBD dataset is publicly available at \href{https://xview2.org/dataset}{https://xview2.org/dataset}. LEVIR-CD dataset can be acquired from \href{https://justchenhao.github.io/LEVIR/}{https://justchenhao.github.io/LEVIR/}. Ida-BD is publicly available on DesignSafe-CI at \href{https://www.designsafe-ci.org/data/browser/public/designsafe.storage.published/PRJ-3563}{https://www.designsafe-ci.org/data/browser/public/designsafe.storage.published/PRJ-3563}.

\bibliography{wileyNJD-APA}%

\begin{thebibliography}{}

\bibitem [\protect \citeauthoryear {%
Alam%
\ \protect \BOthers {.}}{%
Alam%
\ \protect \BOthers {.}}{%
{\protect \APACyear {2020}}%
}]{%
dynamicensemble}
\APACinsertmetastar {%
dynamicensemble}%
\begin{APACrefauthors}%
Alam, K\BPBI M\BPBI R.%
, Siddique, N.%
\BCBL {}\ \BBA {} Adeli, H.%
\end{APACrefauthors}%
\unskip\
\newblock
\APACrefYearMonthDay{2020}{jun}{}.
\newblock
{\BBOQ}\APACrefatitle {A Dynamic Ensemble Learning Algorithm for Neural
  Networks} {A dynamic ensemble learning algorithm for neural networks}.{\BBCQ}
\newblock
\APACjournalVolNumPages{Neural Comput. Appl.}{32}{12}{8675–8690}.
\newblock
\begin{APACrefURL} \url{https://doi.org/10.1007/s00521-019-04359-7}
  \end{APACrefURL}
\newblock
\begin{APACrefDOI} \doi{10.1007/s00521-019-04359-7} \end{APACrefDOI}
\PrintBackRefs{\CurrentBib}

\bibitem [\protect \citeauthoryear {%
Amezquita-Sanchez%
\ \protect \BOthers {.}}{%
Amezquita-Sanchez%
\ \protect \BOthers {.}}{%
{\protect \APACyear {2017}}%
}]{%
amezquita-sanchezNovelMethodologyModal2017}
\APACinsertmetastar {%
amezquita-sanchezNovelMethodologyModal2017}%
\begin{APACrefauthors}%
Amezquita-Sanchez, J\BPBI P.%
, Park, H\BPBI S.%
\BCBL {}\ \BBA {} Adeli, H.%
\end{APACrefauthors}%
\unskip\
\newblock
\APACrefYearMonthDay{2017}{}{}.
\newblock
{\BBOQ}\APACrefatitle {A Novel Methodology for Modal Parameters Identification
  of Large Smart Structures Using {{MUSIC}}, Empirical Wavelet Transform, and
  {{Hilbert}} Transform} {A novel methodology for modal parameters
  identification of large smart structures using {{MUSIC}}, empirical wavelet
  transform, and {{Hilbert}} transform}.{\BBCQ}
\newblock
\APACjournalVolNumPages{Engineering Structures}{147}{}{148--159}.
\newblock
\begin{APACrefURL}
  \url{https://www.sciencedirect.com/science/article/pii/S0141029617300445}
  \end{APACrefURL}
\PrintBackRefs{\CurrentBib}

\bibitem [\protect \citeauthoryear {%
Amezquita-Sanchez%
\ \protect \BOthers {.}}{%
Amezquita-Sanchez%
\ \protect \BOthers {.}}{%
{\protect \APACyear {2018}}%
}]{%
amezquita-sanchezWirelessSmartSensors2018}
\APACinsertmetastar {%
amezquita-sanchezWirelessSmartSensors2018}%
\begin{APACrefauthors}%
Amezquita-Sanchez, J\BPBI P.%
, Valtierra-Rodriguez, M.%
\BCBL {}\ \BBA {} Adeli, H.%
\end{APACrefauthors}%
\unskip\
\newblock
\APACrefYearMonthDay{2018}{}{}.
\newblock
{\BBOQ}\APACrefatitle {Wireless Smart Sensors for Monitoring the Health
  Condition of Civil Infrastructure} {Wireless smart sensors for monitoring the
  health condition of civil infrastructure}.{\BBCQ}
\newblock
\APACjournalVolNumPages{Scientia Iranica}{25}{6}{2913--2925}.
\newblock
\begin{APACrefURL} \url{http://scientiairanica.sharif.edu/article_21136.html}
  \end{APACrefURL}
\PrintBackRefs{\CurrentBib}

\bibitem [\protect \citeauthoryear {%
Athanasiou%
\ \protect \BOthers {.}}{%
Athanasiou%
\ \protect \BOthers {.}}{%
{\protect \APACyear {2020}}%
}]{%
athanasiouMachineLearningApproach2020}
\APACinsertmetastar {%
athanasiouMachineLearningApproach2020}%
\begin{APACrefauthors}%
Athanasiou, A.%
, Ebrahimkhanlou, A.%
, Zaborac, J.%
, Hrynyk, T.%
\BCBL {}\ \BBA {} Salamone, S.%
\end{APACrefauthors}%
\unskip\
\newblock
\APACrefYearMonthDay{2020}{}{}.
\newblock
{\BBOQ}\APACrefatitle {A Machine Learning Approach Based on Multifractal
  Features for Crack Assessment of Reinforced Concrete Shells} {A machine
  learning approach based on multifractal features for crack assessment of
  reinforced concrete shells}.{\BBCQ}
\newblock
\APACjournalVolNumPages{Computer-Aided Civil and Infrastructure
  Engineering}{35}{6}{565--578}.
\newblock
\begin{APACrefURL}
  \url{https://onlinelibrary.wiley.com/doi/abs/10.1111/mice.12509}
  \end{APACrefURL}
\PrintBackRefs{\CurrentBib}

\bibitem [\protect \citeauthoryear {%
Bandara%
\ \BBA {} Patel%
}{%
Bandara%
\ \BBA {} Patel%
}{%
{\protect \APACyear {2022}}%
}]{%
changeformer}
\APACinsertmetastar {%
changeformer}%
\begin{APACrefauthors}%
Bandara, W\BPBI G\BPBI C.%
\BCBT {}\ \BBA {} Patel, V\BPBI M.%
\end{APACrefauthors}%
\unskip\
\newblock
\APACrefYearMonthDay{2022}{}{}.
\newblock
\APACrefbtitle {A Transformer-Based Siamese Network for Change Detection.} {A
  transformer-based siamese network for change detection.}
\newblock
\APACaddressPublisher{}{arXiv}.
\newblock
\begin{APACrefURL} \url{https://arxiv.org/abs/2201.01293} \end{APACrefURL}
\PrintBackRefs{\CurrentBib}

\bibitem [\protect \citeauthoryear {%
Beven~II%
\ \protect \BOthers {.}}{%
Beven~II%
\ \protect \BOthers {.}}{%
{\protect \APACyear {2022}}%
}]{%
beven_ii_national_2022}
\APACinsertmetastar {%
beven_ii_national_2022}%
\begin{APACrefauthors}%
Beven~II, J\BPBI L.%
, Hagen, A.%
\BCBL {}\ \BBA {} Berg, R.%
\end{APACrefauthors}%
\unskip\
\newblock
\APACrefYearMonthDay{2022}{{\APACmonth{04}}}{}.
\newblock
\APACrefbtitle {National {Hurricane} {Center} {Tropical} {Cyclone} {Report} -
  {Hurrican} {Ida}} {National {Hurricane} {Center} {Tropical} {Cyclone}
  {Report} - {Hurrican} {Ida}}\ \APACbVolEdTR{}{\BTR{}\ \BNUM\ AL092021}.
\newblock
\APACaddressInstitution{}{Nationa Hurricane Center}.
\newblock
\begin{APACrefURL} [{2022-07-08}]
  \url{https://www.nhc.noaa.gov/data/tcr/AL092021_Ida.pdf} \end{APACrefURL}
\PrintBackRefs{\CurrentBib}

\bibitem [\protect \citeauthoryear {%
Cao%
\ \BBA {} Choe%
}{%
Cao%
\ \BBA {} Choe%
}{%
{\protect \APACyear {2020}}%
}]{%
cao_building_2020}
\APACinsertmetastar {%
cao_building_2020}%
\begin{APACrefauthors}%
Cao, Q\BPBI D.%
\BCBT {}\ \BBA {} Choe, Y.%
\end{APACrefauthors}%
\unskip\
\newblock
\APACrefYearMonthDay{2020}{{\APACmonth{09}}}{}.
\newblock
{\BBOQ}\APACrefatitle {Building damage annotation on post-hurricane satellite
  imagery based on convolutional neural networks} {Building damage annotation
  on post-hurricane satellite imagery based on convolutional neural
  networks}.{\BBCQ}
\newblock
\APACjournalVolNumPages{Natural Hazards}{103}{3}{3357--3376}.
\PrintBackRefs{\CurrentBib}

\bibitem [\protect \citeauthoryear {%
Chen%
\ \protect \BOthers {.}}{%
Chen%
\ \protect \BOthers {.}}{%
{\protect \APACyear {2022}}%
}]{%
chen2021bit}
\APACinsertmetastar {%
chen2021bit}%
\begin{APACrefauthors}%
Chen, H.%
, Qi, Z.%
\BCBL {}\ \BBA {} Shi, Z.%
\end{APACrefauthors}%
\unskip\
\newblock
\APACrefYearMonthDay{2022}{}{}.
\newblock
{\BBOQ}\APACrefatitle {Remote Sensing Image Change Detection With Transformers}
  {Remote sensing image change detection with transformers}.{\BBCQ}
\newblock
\APACjournalVolNumPages{IEEE Transactions on Geoscience and Remote
  Sensing}{60}{}{1-14}.
\PrintBackRefs{\CurrentBib}

\bibitem [\protect \citeauthoryear {%
Cheng%
\ \protect \BOthers {.}}{%
Cheng%
\ \protect \BOthers {.}}{%
{\protect \APACyear {2021}}%
}]{%
cheng_deep_2021}
\APACinsertmetastar {%
cheng_deep_2021}%
\begin{APACrefauthors}%
Cheng, C.%
, Behzadan, A\BPBI H.%
\BCBL {}\ \BBA {} Noshadravan, A.%
\end{APACrefauthors}%
\unskip\
\newblock
\APACrefYearMonthDay{2021}{{\APACmonth{02}}}{}.
\newblock
{\BBOQ}\APACrefatitle {Deep learning for post‐hurricane aerial damage
  assessment of buildings} {Deep learning for post‐hurricane aerial damage
  assessment of buildings}.{\BBCQ}
\newblock
\APACjournalVolNumPages{Computer-Aided Civil and Infrastructure
  Engineering}{}{}{mice.12658}.
\newblock
\begin{APACrefURL} [{2021-05-20}]
  \url{https://onlinelibrary.wiley.com/doi/10.1111/mice.12658} \end{APACrefURL}
\PrintBackRefs{\CurrentBib}

\bibitem [\protect \citeauthoryear {%
Corbane%
\ \protect \BOthers {.}}{%
Corbane%
\ \protect \BOthers {.}}{%
{\protect \APACyear {2011}}%
}]{%
corbane_comparison_2011}
\APACinsertmetastar {%
corbane_comparison_2011}%
\begin{APACrefauthors}%
Corbane, C.%
, Carrion, D.%
, Lemoine, G.%
\BCBL {}\ \BBA {} Broglia, M.%
\end{APACrefauthors}%
\unskip\
\newblock
\APACrefYearMonthDay{2011}{{\APACmonth{10}}}{}.
\newblock
{\BBOQ}\APACrefatitle {Comparison of {Damage} {Assessment} {Maps} {Derived}
  from {Very} {High} {Spatial} {Resolution} {Satellite} and {Aerial} {Imagery}
  {Produced} for the {Haiti} 2010 {Earthquake}} {Comparison of {Damage}
  {Assessment} {Maps} {Derived} from {Very} {High} {Spatial} {Resolution}
  {Satellite} and {Aerial} {Imagery} {Produced} for the {Haiti} 2010
  {Earthquake}}.{\BBCQ}
\newblock
\APACjournalVolNumPages{Earthquake Spectra}{27}{1\_suppl1}{199--218}.
\PrintBackRefs{\CurrentBib}

\bibitem [\protect \citeauthoryear {%
Fujita%
\ \protect \BOthers {.}}{%
Fujita%
\ \protect \BOthers {.}}{%
{\protect \APACyear {2017}}%
}]{%
fujita_damage_2017}
\APACinsertmetastar {%
fujita_damage_2017}%
\begin{APACrefauthors}%
Fujita, A.%
, Sakurada, K.%
, Imaizumi, T.%
, Ito, R.%
, Hikosaka, S.%
\BCBL {}\ \BBA {} Nakamura, R.%
\end{APACrefauthors}%
\unskip\
\newblock
\APACrefYearMonthDay{2017}{{\APACmonth{05}}}{}.
\newblock
{\BBOQ}\APACrefatitle {Damage detection from aerial images via convolutional
  neural networks} {Damage detection from aerial images via convolutional
  neural networks}.{\BBCQ}
\newblock
\BIn{} \APACrefbtitle {2017 {Fifteenth} {IAPR} {International} {Conference} on
  {Machine} {Vision} {Applications} ({MVA})} {2017 {Fifteenth} {IAPR}
  {International} {Conference} on {Machine} {Vision} {Applications} ({MVA})}\
  (\BPGS\ 5--8).
\PrintBackRefs{\CurrentBib}

\bibitem [\protect \citeauthoryear {%
Gao%
\ \BBA {} Mosalam%
}{%
Gao%
\ \BBA {} Mosalam%
}{%
{\protect \APACyear {2018}}%
}]{%
gaoDeepTransferLearning2018}
\APACinsertmetastar {%
gaoDeepTransferLearning2018}%
\begin{APACrefauthors}%
Gao, Y.%
\BCBT {}\ \BBA {} Mosalam, K\BPBI M.%
\end{APACrefauthors}%
\unskip\
\newblock
\APACrefYearMonthDay{2018}{}{}.
\newblock
{\BBOQ}\APACrefatitle {Deep {{Transfer Learning}} for {{Image-Based Structural
  Damage Recognition}}} {Deep {{Transfer Learning}} for {{Image-Based
  Structural Damage Recognition}}}.{\BBCQ}
\newblock
\APACjournalVolNumPages{Computer-Aided Civil and Infrastructure
  Engineering}{33}{9}{748--768}.
\newblock
\begin{APACrefURL}
  \url{https://onlinelibrary.wiley.com/doi/abs/10.1111/mice.12363}
  \end{APACrefURL}
\PrintBackRefs{\CurrentBib}

\bibitem [\protect \citeauthoryear {%
Guo%
\ \protect \BOthers {.}}{%
Guo%
\ \protect \BOthers {.}}{%
{\protect \APACyear {2020}}%
}]{%
few-shot-learning}
\APACinsertmetastar {%
few-shot-learning}%
\begin{APACrefauthors}%
Guo, Y.%
, Codella, N\BPBI C.%
, Karlinsky, L.%
, Codella, J\BPBI V.%
, Smith, J\BPBI R.%
, Saenko, K.%
\BDBL {}Feris, R.%
\end{APACrefauthors}%
\unskip\
\newblock
\APACrefYearMonthDay{2020}{}{}.
\newblock
{\BBOQ}\APACrefatitle {A Broader Study of Cross-Domain Few-Shot Learning} {A
  broader study of cross-domain few-shot learning}.{\BBCQ}
\newblock
\BIn{} A.~Vedaldi, H.~Bischof, T.~Brox\BCBL {}\ \BBA {} J\BHBI M.~Frahm\
  (\BEDS), \APACrefbtitle {Computer Vision -- ECCV 2020} {Computer vision --
  eccv 2020}\ (\BPGS\ 124--141).
\newblock
\APACaddressPublisher{Cham}{Springer International Publishing}.
\PrintBackRefs{\CurrentBib}

\bibitem [\protect \citeauthoryear {%
Gupta%
\ \protect \BOthers {.}}{%
Gupta%
\ \protect \BOthers {.}}{%
{\protect \APACyear {2019}}%
}]{%
xBDdataset}
\APACinsertmetastar {%
xBDdataset}%
\begin{APACrefauthors}%
Gupta, R.%
, Hosfelt, R.%
, Sajeev, S.%
, Patel, N.%
, Goodman, B.%
, Doshi, J.%
\BDBL {}Gaston, M\BPBI E.%
\end{APACrefauthors}%
\unskip\
\newblock
\APACrefYearMonthDay{2019}{}{}.
\newblock
{\BBOQ}\APACrefatitle {{xBD}: {A} Dataset for Assessing Building Damage from
  Satellite Imagery} {{xBD}: {A} dataset for assessing building damage from
  satellite imagery}.{\BBCQ}
\newblock
\APACjournalVolNumPages{CoRR}{abs/1911.09296}{}{}.
\newblock
\begin{APACrefURL} \url{http://arxiv.org/abs/1911.09296} \end{APACrefURL}
\PrintBackRefs{\CurrentBib}

\bibitem [\protect \citeauthoryear {%
Gupta%
\ \BBA {} Shah%
}{%
Gupta%
\ \BBA {} Shah%
}{%
{\protect \APACyear {2021}}%
}]{%
rescuenet}
\APACinsertmetastar {%
rescuenet}%
\begin{APACrefauthors}%
Gupta, R.%
\BCBT {}\ \BBA {} Shah, M.%
\end{APACrefauthors}%
\unskip\
\newblock
\APACrefYearMonthDay{2021}{}{}.
\newblock
{\BBOQ}\APACrefatitle {RescueNet: Joint Building Segmentation and Damage
  Assessment from Satellite Imagery} {Rescuenet: Joint building segmentation
  and damage assessment from satellite imagery}.{\BBCQ}
\newblock
\BIn{} \APACrefbtitle {2020 25th International Conference on Pattern
  Recognition (ICPR)} {2020 25th international conference on pattern
  recognition (icpr)}\ (\BPG~4405-4411).
\PrintBackRefs{\CurrentBib}

\bibitem [\protect \citeauthoryear {%
Hao%
\ \protect \BOthers {.}}{%
Hao%
\ \protect \BOthers {.}}{%
{\protect \APACyear {2021}}%
}]{%
siam-att-unet2}
\APACinsertmetastar {%
siam-att-unet2}%
\begin{APACrefauthors}%
Hao, H.%
, Baireddy, S.%
, Bartusiak, E\BPBI R.%
, Konz, L.%
, LaTourette, K.%
, Gribbons, M.%
\BDBL {}Comer, M\BPBI L.%
\end{APACrefauthors}%
\unskip\
\newblock
\APACrefYearMonthDay{2021}{}{}.
\newblock
{\BBOQ}\APACrefatitle {An Attention-Based System for Damage Assessment Using
  Satellite Imagery} {An attention-based system for damage assessment using
  satellite imagery}.{\BBCQ}
\newblock
\BIn{} \APACrefbtitle {2021 IEEE International Geoscience and Remote Sensing
  Symposium IGARSS} {2021 ieee international geoscience and remote sensing
  symposium igarss}\ (\BPG~4396-4399).
\PrintBackRefs{\CurrentBib}

\bibitem [\protect \citeauthoryear {%
Hinton%
\ \protect \BOthers {.}}{%
Hinton%
\ \protect \BOthers {.}}{%
{\protect \APACyear {2015}}%
}]{%
knowledge_distill}
\APACinsertmetastar {%
knowledge_distill}%
\begin{APACrefauthors}%
Hinton, G.%
, Vinyals, O.%
\BCBL {}\ \BBA {} Dean, J.%
\end{APACrefauthors}%
\unskip\
\newblock
\APACrefYearMonthDay{2015}{}{}.
\newblock
\APACrefbtitle {Distilling the Knowledge in a Neural Network.} {Distilling the
  knowledge in a neural network.}
\newblock
\APACaddressPublisher{}{arXiv}.
\newblock
\begin{APACrefURL} \url{https://arxiv.org/abs/1503.02531} \end{APACrefURL}
\newblock
\begin{APACrefDOI} \doi{10.48550/ARXIV.1503.02531} \end{APACrefDOI}
\PrintBackRefs{\CurrentBib}

\bibitem [\protect \citeauthoryear {%
Jr%
\ \BBA {} Adeli%
}{%
Jr%
\ \BBA {} Adeli%
}{%
{\protect \APACyear {2018}}%
}]{%
jrInfraredThermographyDetecting2018}
\APACinsertmetastar {%
jrInfraredThermographyDetecting2018}%
\begin{APACrefauthors}%
Jr, G\BPBI F\BPBI S.%
\BCBT {}\ \BBA {} Adeli, H.%
\end{APACrefauthors}%
\unskip\
\newblock
\APACrefYearMonthDay{2018}{}{}.
\newblock
{\BBOQ}\APACrefatitle {Infrared Thermography for Detecting Defects in Concrete
  Structures} {Infrared thermography for detecting defects in concrete
  structures}.{\BBCQ}
\newblock
\APACjournalVolNumPages{Journal of Civil Engineering and
  Management}{24}{7}{508--515}.
\newblock
\begin{APACrefURL}
  \url{https://journals.vilniustech.lt/index.php/JCEM/article/view/6186}
  \end{APACrefURL}
\PrintBackRefs{\CurrentBib}

\bibitem [\protect \citeauthoryear {%
Khajwal%
\ \protect \BOthers {.}}{%
Khajwal%
\ \protect \BOthers {.}}{%
{\protect \APACyear {2022}}%
}]{%
khajwalPostdisasterDamageClassification2022}
\APACinsertmetastar {%
khajwalPostdisasterDamageClassification2022}%
\begin{APACrefauthors}%
Khajwal, A\BPBI B.%
, Cheng, C\BHBI S.%
\BCBL {}\ \BBA {} Noshadravan, A.%
\end{APACrefauthors}%
\unskip\
\newblock
\APACrefYearMonthDay{2022}{}{}.
\newblock
{\BBOQ}\APACrefatitle {Post-Disaster Damage Classification Based on Deep
  Multi-View Image Fusion} {Post-disaster damage classification based on deep
  multi-view image fusion}.{\BBCQ}
\newblock
\APACjournalVolNumPages{Computer-Aided Civil and Infrastructure
  Engineering}{n/a}{n/a}{}.
\newblock
\begin{APACrefURL}
  \url{https://onlinelibrary.wiley.com/doi/abs/10.1111/mice.12890}
  \end{APACrefURL}
\PrintBackRefs{\CurrentBib}

\bibitem [\protect \citeauthoryear {%
Ku%
\ \protect \BOthers {.}}{%
Ku%
\ \protect \BOthers {.}}{%
{\protect \APACyear {2020}}%
}]{%
dualhrnet}
\APACinsertmetastar {%
dualhrnet}%
\begin{APACrefauthors}%
Ku, J.%
, Seo, J.%
\BCBL {}\ \BBA {} Jeon, T.%
\end{APACrefauthors}%
\unskip\
\newblock
\APACrefYearMonthDay{2020}{}{}.
\newblock
\APACrefbtitle {DualHRNet for Building Localization and DamageClassification.}
  {Dualhrnet for building localization and damageclassification.}
\newblock
\begin{APACrefURL}
  \url{https://github.com/SIAnalytics/dual-hrnet/blob/master/figures/xView2_White_Paper_SI_Analytics.pdf}
  \end{APACrefURL}
\PrintBackRefs{\CurrentBib}

\bibitem [\protect \citeauthoryear {%
{Labelbox}%
}{%
{Labelbox}%
}{%
{\protect \APACyear {2022}}%
}]{%
labelbox_labelbox_2022}
\APACinsertmetastar {%
labelbox_labelbox_2022}%
\begin{APACrefauthors}%
{Labelbox}.%
\end{APACrefauthors}%
\unskip\
\newblock
\APACrefYearMonthDay{2022}{}{}.
\newblock
\APACrefbtitle {Labelbox.} {Labelbox.}
\newblock
\begin{APACrefURL} [{2022-05-18}] \url{https://labelbox.com} \end{APACrefURL}
\PrintBackRefs{\CurrentBib}

\bibitem [\protect \citeauthoryear {%
Lee%
\ \protect \BOthers {.}}{%
Lee%
\ \protect \BOthers {.}}{%
{\protect \APACyear {2022}}%
}]{%
lee_ida-bd_2022}
\APACinsertmetastar {%
lee_ida-bd_2022}%
\begin{APACrefauthors}%
Lee, C\BHBI C.%
, Kaur, N.%
, Mahdavi-Amiri, A.%
\BCBL {}\ \BBA {} Mostafavi, A.%
\end{APACrefauthors}%
\unskip\
\newblock
\APACrefYearMonthDay{2022}{}{}.
\newblock
\APACrefbtitle {Ida-{BD}: pre- and post-disaster high-resolution satellite
  imagery for building damage assessment from {Hurricane} {Ida}.} {Ida-{BD}:
  pre- and post-disaster high-resolution satellite imagery for building damage
  assessment from {Hurricane} {Ida}.}
\newblock
\APACaddressPublisher{}{Designsafe-CI}.
\newblock
\begin{APACrefURL} [{2022-07-25}]
  \url{https://www.designsafe-ci.org/data/browser/public/designsafe.storage.published/PRJ-3563}
  \end{APACrefURL}
\PrintBackRefs{\CurrentBib}

\bibitem [\protect \citeauthoryear {%
S.~Li%
\ \protect \BOthers {.}}{%
S.~Li%
\ \protect \BOthers {.}}{%
{\protect \APACyear {2019}}%
}]{%
liAutomaticPixellevelMultiple2019}
\APACinsertmetastar {%
liAutomaticPixellevelMultiple2019}%
\begin{APACrefauthors}%
Li, S.%
, Zhao, X.%
\BCBL {}\ \BBA {} Zhou, G.%
\end{APACrefauthors}%
\unskip\
\newblock
\APACrefYearMonthDay{2019}{}{}.
\newblock
{\BBOQ}\APACrefatitle {Automatic Pixel-Level Multiple Damage Detection of
  Concrete Structure Using Fully Convolutional Network} {Automatic pixel-level
  multiple damage detection of concrete structure using fully convolutional
  network}.{\BBCQ}
\newblock
\APACjournalVolNumPages{Computer-Aided Civil and Infrastructure
  Engineering}{34}{7}{616--634}.
\newblock
\begin{APACrefURL}
  \url{https://onlinelibrary.wiley.com/doi/abs/10.1111/mice.12433}
  \end{APACrefURL}
\PrintBackRefs{\CurrentBib}

\bibitem [\protect \citeauthoryear {%
Z.~Li%
\ \protect \BOthers {.}}{%
Z.~Li%
\ \protect \BOthers {.}}{%
{\protect \APACyear {2017}}%
}]{%
liNewMethodModal2017}
\APACinsertmetastar {%
liNewMethodModal2017}%
\begin{APACrefauthors}%
Li, Z.%
, Park, H\BPBI S.%
\BCBL {}\ \BBA {} Adeli, H.%
\end{APACrefauthors}%
\unskip\
\newblock
\APACrefYearMonthDay{2017}{}{}.
\newblock
{\BBOQ}\APACrefatitle {New Method for Modal Identification of Super High-Rise
  Building Structures Using Discretized Synchrosqueezed Wavelet and {{Hilbert}}
  Transforms} {New method for modal identification of super high-rise building
  structures using discretized synchrosqueezed wavelet and {{Hilbert}}
  transforms}.{\BBCQ}
\newblock
\APACjournalVolNumPages{The Structural Design of Tall and Special
  Buildings}{26}{3}{e1312}.
\newblock
\begin{APACrefURL}
  \url{https://onlinelibrary.wiley.com/doi/abs/10.1002/tal.1312}
  \end{APACrefURL}
\PrintBackRefs{\CurrentBib}

\bibitem [\protect \citeauthoryear {%
Lin%
\ \protect \BOthers {.}}{%
Lin%
\ \protect \BOthers {.}}{%
{\protect \APACyear {2017}}%
}]{%
focal_loss}
\APACinsertmetastar {%
focal_loss}%
\begin{APACrefauthors}%
Lin, T\BHBI Y.%
, Goyal, P.%
, Girshick, R.%
, He, K.%
\BCBL {}\ \BBA {} Dollar, P.%
\end{APACrefauthors}%
\unskip\
\newblock
\APACrefYearMonthDay{2017}{Oct}{}.
\newblock
{\BBOQ}\APACrefatitle {Focal Loss for Dense Object Detection} {Focal loss for
  dense object detection}.{\BBCQ}
\newblock
\BIn{} \APACrefbtitle {Proceedings of the IEEE International Conference on
  Computer Vision (ICCV).} {Proceedings of the ieee international conference on
  computer vision (iccv).}
\PrintBackRefs{\CurrentBib}

\bibitem [\protect \citeauthoryear {%
Liu%
\ \protect \BOthers {.}}{%
Liu%
\ \protect \BOthers {.}}{%
{\protect \APACyear {2022}}%
}]{%
cropland}
\APACinsertmetastar {%
cropland}%
\begin{APACrefauthors}%
Liu, M.%
, Chai, Z.%
, Deng, H.%
\BCBL {}\ \BBA {} Liu, R.%
\end{APACrefauthors}%
\unskip\
\newblock
\APACrefYearMonthDay{2022}{}{}.
\newblock
{\BBOQ}\APACrefatitle {A CNN-Transformer Network With Multiscale Context
  Aggregation for Fine-Grained Cropland Change Detection} {A cnn-transformer
  network with multiscale context aggregation for fine-grained cropland change
  detection}.{\BBCQ}
\newblock
\APACjournalVolNumPages{IEEE Journal of Selected Topics in Applied Earth
  Observations and Remote Sensing}{15}{}{4297-4306}.
\newblock
\begin{APACrefDOI} \doi{10.1109/JSTARS.2022.3177235} \end{APACrefDOI}
\PrintBackRefs{\CurrentBib}

\bibitem [\protect \citeauthoryear {%
McCarthy%
\ \protect \BOthers {.}}{%
McCarthy%
\ \protect \BOthers {.}}{%
{\protect \APACyear {2020}}%
}]{%
mccarthy_mapping_2020}
\APACinsertmetastar {%
mccarthy_mapping_2020}%
\begin{APACrefauthors}%
McCarthy, M\BPBI J.%
, Jessen, B.%
, Barry, M\BPBI J.%
, Figueroa, M.%
, McIntosh, J.%
, Murray, T.%
\BDBL {}Muller-Karger, F\BPBI E.%
\end{APACrefauthors}%
\unskip\
\newblock
\APACrefYearMonthDay{2020}{{\APACmonth{09}}}{}.
\newblock
{\BBOQ}\APACrefatitle {Mapping hurricane damage: {A} comparative analysis of
  satellite monitoring methods} {Mapping hurricane damage: {A} comparative
  analysis of satellite monitoring methods}.{\BBCQ}
\newblock
\APACjournalVolNumPages{International Journal of Applied Earth Observation and
  Geoinformation}{91}{}{102134}.
\PrintBackRefs{\CurrentBib}

\bibitem [\protect \citeauthoryear {%
Milletari%
\ \protect \BOthers {.}}{%
Milletari%
\ \protect \BOthers {.}}{%
{\protect \APACyear {2016}}%
}]{%
dice_loss}
\APACinsertmetastar {%
dice_loss}%
\begin{APACrefauthors}%
Milletari, F.%
, Navab, N.%
\BCBL {}\ \BBA {} Ahmadi, S\BHBI A.%
\end{APACrefauthors}%
\unskip\
\newblock
\APACrefYearMonthDay{2016}{}{}.
\newblock
{\BBOQ}\APACrefatitle {V-Net: Fully Convolutional Neural Networks for
  Volumetric Medical Image Segmentation} {V-net: Fully convolutional neural
  networks for volumetric medical image segmentation}.{\BBCQ}
\newblock
\BIn{} \APACrefbtitle {2016 Fourth International Conference on 3D Vision (3DV)}
  {2016 fourth international conference on 3d vision (3dv)}\ (\BPG~565-571).
\PrintBackRefs{\CurrentBib}

\bibitem [\protect \citeauthoryear {%
Oh%
\ \protect \BOthers {.}}{%
Oh%
\ \protect \BOthers {.}}{%
{\protect \APACyear {2017}}%
}]{%
ohEvolutionaryLearningBased2017}
\APACinsertmetastar {%
ohEvolutionaryLearningBased2017}%
\begin{APACrefauthors}%
Oh, B\BPBI K.%
, Kim, K\BPBI J.%
, Kim, Y.%
, Park, H\BPBI S.%
\BCBL {}\ \BBA {} Adeli, H.%
\end{APACrefauthors}%
\unskip\
\newblock
\APACrefYearMonthDay{2017}{}{}.
\newblock
{\BBOQ}\APACrefatitle {Evolutionary Learning Based Sustainable Strain Sensing
  Model for Structural Health Monitoring of High-Rise Buildings} {Evolutionary
  learning based sustainable strain sensing model for structural health
  monitoring of high-rise buildings}.{\BBCQ}
\newblock
\APACjournalVolNumPages{Applied Soft Computing}{58}{}{576--585}.
\newblock
\begin{APACrefURL}
  \url{https://www.sciencedirect.com/science/article/pii/S1568494617302922}
  \end{APACrefURL}
\PrintBackRefs{\CurrentBib}

\bibitem [\protect \citeauthoryear {%
Pereira%
\ \protect \BOthers {.}}{%
Pereira%
\ \protect \BOthers {.}}{%
{\protect \APACyear {2020}}%
}]{%
fema}
\APACinsertmetastar {%
fema}%
\begin{APACrefauthors}%
Pereira, D\BPBI R.%
, Piteri, M\BPBI A.%
, Souza, A\BPBI N.%
, Papa, J\BPBI a\BPBI P.%
\BCBL {}\ \BBA {} Adeli, H.%
\end{APACrefauthors}%
\unskip\
\newblock
\APACrefYearMonthDay{2020}{may}{}.
\newblock
{\BBOQ}\APACrefatitle {FEMa: A Finite Element Machine for Fast Learning} {Fema:
  A finite element machine for fast learning}.{\BBCQ}
\newblock
\APACjournalVolNumPages{Neural Comput. Appl.}{32}{10}{6393–6404}.
\newblock
\begin{APACrefURL} \url{https://doi.org/10.1007/s00521-019-04146-4}
  \end{APACrefURL}
\newblock
\begin{APACrefDOI} \doi{10.1007/s00521-019-04146-4} \end{APACrefDOI}
\PrintBackRefs{\CurrentBib}

\bibitem [\protect \citeauthoryear {%
Perez-Ramirez%
\ \protect \BOthers {.}}{%
Perez-Ramirez%
\ \protect \BOthers {.}}{%
{\protect \APACyear {2019}}%
}]{%
perez-ramirezRecurrentNeuralNetwork2019}
\APACinsertmetastar {%
perez-ramirezRecurrentNeuralNetwork2019}%
\begin{APACrefauthors}%
Perez-Ramirez, C\BPBI A.%
, Amezquita-Sanchez, J\BPBI P.%
, Valtierra-Rodriguez, M.%
, Adeli, H.%
, Dominguez-Gonzalez, A.%
\BCBL {}\ \BBA {} Romero-Troncoso, R\BPBI J.%
\end{APACrefauthors}%
\unskip\
\newblock
\APACrefYearMonthDay{2019}{}{}.
\newblock
{\BBOQ}\APACrefatitle {Recurrent Neural Network Model with {{Bayesian}}
  Training and Mutual Information for Response Prediction of Large Buildings}
  {Recurrent neural network model with {{Bayesian}} training and mutual
  information for response prediction of large buildings}.{\BBCQ}
\newblock
\APACjournalVolNumPages{Engineering Structures}{178}{}{603--615}.
\newblock
\begin{APACrefURL}
  \url{https://www.sciencedirect.com/science/article/pii/S0141029618307235}
  \end{APACrefURL}
\PrintBackRefs{\CurrentBib}

\bibitem [\protect \citeauthoryear {%
Rafiei%
\ \BBA {} Adeli%
}{%
Rafiei%
\ \BBA {} Adeli%
}{%
{\protect \APACyear {2017}}%
}]{%
neuraladelhi}
\APACinsertmetastar {%
neuraladelhi}%
\begin{APACrefauthors}%
Rafiei, M\BPBI H.%
\BCBT {}\ \BBA {} Adeli, H.%
\end{APACrefauthors}%
\unskip\
\newblock
\APACrefYearMonthDay{2017}{}{}.
\newblock
{\BBOQ}\APACrefatitle {A New Neural Dynamic Classification Algorithm} {A new
  neural dynamic classification algorithm}.{\BBCQ}
\newblock
\APACjournalVolNumPages{IEEE Transactions on Neural Networks and Learning
  Systems}{28}{12}{3074-3083}.
\newblock
\begin{APACrefDOI} \doi{10.1109/TNNLS.2017.2682102} \end{APACrefDOI}
\PrintBackRefs{\CurrentBib}

\bibitem [\protect \citeauthoryear {%
Ronneberger%
\ \protect \BOthers {.}}{%
Ronneberger%
\ \protect \BOthers {.}}{%
{\protect \APACyear {2015}}%
}]{%
unet}
\APACinsertmetastar {%
unet}%
\begin{APACrefauthors}%
Ronneberger, O.%
, Fischer, P.%
\BCBL {}\ \BBA {} Brox, T.%
\end{APACrefauthors}%
\unskip\
\newblock
\APACrefYearMonthDay{2015}{}{}.
\newblock
{\BBOQ}\APACrefatitle {U-Net: Convolutional Networks for Biomedical Image
  Segmentation} {U-net: Convolutional networks for biomedical image
  segmentation}.{\BBCQ}
\newblock
\APACjournalVolNumPages{CoRR}{abs/1505.04597}{}{}.
\newblock
\begin{APACrefURL} \url{http://arxiv.org/abs/1505.04597} \end{APACrefURL}
\PrintBackRefs{\CurrentBib}

\bibitem [\protect \citeauthoryear {%
Sajedi%
\ \BBA {} Liang%
}{%
Sajedi%
\ \BBA {} Liang%
}{%
{\protect \APACyear {2020}}%
}]{%
sajediVibrationbasedSemanticDamage2020}
\APACinsertmetastar {%
sajediVibrationbasedSemanticDamage2020}%
\begin{APACrefauthors}%
Sajedi, S\BPBI O.%
\BCBT {}\ \BBA {} Liang, X.%
\end{APACrefauthors}%
\unskip\
\newblock
\APACrefYearMonthDay{2020}{}{}.
\newblock
{\BBOQ}\APACrefatitle {Vibration-Based Semantic Damage Segmentation for
  Large-Scale Structural Health Monitoring} {Vibration-based semantic damage
  segmentation for large-scale structural health monitoring}.{\BBCQ}
\newblock
\APACjournalVolNumPages{Computer-Aided Civil and Infrastructure
  Engineering}{35}{6}{579--596}.
\newblock
\begin{APACrefURL}
  \url{https://onlinelibrary.wiley.com/doi/abs/10.1111/mice.12523}
  \end{APACrefURL}
\PrintBackRefs{\CurrentBib}

\bibitem [\protect \citeauthoryear {%
Shen%
\ \protect \BOthers {.}}{%
Shen%
\ \protect \BOthers {.}}{%
{\protect \APACyear {2022}}%
}]{%
bdanet}
\APACinsertmetastar {%
bdanet}%
\begin{APACrefauthors}%
Shen, Y.%
, Zhu, S.%
, Yang, T.%
, Chen, C.%
, Pan, D.%
, Chen, J.%
\BDBL {}Du, Q.%
\end{APACrefauthors}%
\unskip\
\newblock
\APACrefYearMonthDay{2022}{}{}.
\newblock
{\BBOQ}\APACrefatitle {BDANet: Multiscale Convolutional Neural Network With
  Cross-Directional Attention for Building Damage Assessment From Satellite
  Images} {Bdanet: Multiscale convolutional neural network with
  cross-directional attention for building damage assessment from satellite
  images}.{\BBCQ}
\newblock
\APACjournalVolNumPages{IEEE Transactions on Geoscience and Remote
  Sensing}{60}{}{1-14}.
\PrintBackRefs{\CurrentBib}

\bibitem [\protect \citeauthoryear {%
Spencer%
\ \protect \BOthers {.}}{%
Spencer%
\ \protect \BOthers {.}}{%
{\protect \APACyear {2019}}%
}]{%
spencer_advances_2019}
\APACinsertmetastar {%
spencer_advances_2019}%
\begin{APACrefauthors}%
Spencer, B\BPBI F.%
, Hoskere, V.%
\BCBL {}\ \BBA {} Narazaki, Y.%
\end{APACrefauthors}%
\unskip\
\newblock
\APACrefYearMonthDay{2019}{{\APACmonth{04}}}{}.
\newblock
{\BBOQ}\APACrefatitle {Advances in {Computer} {Vision}-{Based} {Civil}
  {Infrastructure} {Inspection} and {Monitoring}} {Advances in {Computer}
  {Vision}-{Based} {Civil} {Infrastructure} {Inspection} and
  {Monitoring}}.{\BBCQ}
\newblock
\APACjournalVolNumPages{Engineering}{5}{2}{199--222}.
\PrintBackRefs{\CurrentBib}

\bibitem [\protect \citeauthoryear {%
Talebinejad%
\ \protect \BOthers {.}}{%
Talebinejad%
\ \protect \BOthers {.}}{%
{\protect \APACyear {2011}}%
}]{%
talebinejadNumericalEvaluationVibrationBased2011}
\APACinsertmetastar {%
talebinejadNumericalEvaluationVibrationBased2011}%
\begin{APACrefauthors}%
Talebinejad, I.%
, Fischer, C.%
\BCBL {}\ \BBA {} Ansari, F.%
\end{APACrefauthors}%
\unskip\
\newblock
\APACrefYearMonthDay{2011}{}{}.
\newblock
{\BBOQ}\APACrefatitle {Numerical {{Evaluation}} of {{Vibration-Based Methods}}
  for {{Damage Assessment}} of {{Cable-Stayed Bridges}}} {Numerical
  {{Evaluation}} of {{Vibration-Based Methods}} for {{Damage Assessment}} of
  {{Cable-Stayed Bridges}}}.{\BBCQ}
\newblock
\APACjournalVolNumPages{Computer-Aided Civil and Infrastructure
  Engineering}{26}{3}{239--251}.
\newblock
\begin{APACrefURL}
  \url{https://onlinelibrary.wiley.com/doi/abs/10.1111/j.1467-8667.2010.00684.x}
  \end{APACrefURL}
\PrintBackRefs{\CurrentBib}

\bibitem [\protect \citeauthoryear {%
Tong%
\ \protect \BOthers {.}}{%
Tong%
\ \protect \BOthers {.}}{%
{\protect \APACyear {2012}}%
}]{%
tong_building-damage_2012}
\APACinsertmetastar {%
tong_building-damage_2012}%
\begin{APACrefauthors}%
Tong, X.%
, Hong, Z.%
, Liu, S.%
, Zhang, X.%
, Xie, H.%
, Li, Z.%
\BDBL {}Bao, F.%
\end{APACrefauthors}%
\unskip\
\newblock
\APACrefYearMonthDay{2012}{{\APACmonth{03}}}{}.
\newblock
{\BBOQ}\APACrefatitle {Building-damage detection using pre- and post-seismic
  high-resolution satellite stereo imagery: {A} case study of the {May} 2008
  {Wenchuan} earthquake} {Building-damage detection using pre- and post-seismic
  high-resolution satellite stereo imagery: {A} case study of the {May} 2008
  {Wenchuan} earthquake}.{\BBCQ}
\newblock
\APACjournalVolNumPages{ISPRS Journal of Photogrammetry and Remote
  Sensing}{68}{}{13--27}.
\PrintBackRefs{\CurrentBib}

\bibitem [\protect \citeauthoryear {%
Tzeng%
\ \protect \BOthers {.}}{%
Tzeng%
\ \protect \BOthers {.}}{%
{\protect \APACyear {2017}}%
}]{%
GAN_domain_adapt}
\APACinsertmetastar {%
GAN_domain_adapt}%
\begin{APACrefauthors}%
Tzeng, E.%
, Hoffman, J.%
, Saenko, K.%
\BCBL {}\ \BBA {} Darrell, T.%
\end{APACrefauthors}%
\unskip\
\newblock
\APACrefYearMonthDay{2017}{July}{}.
\newblock
{\BBOQ}\APACrefatitle {Adversarial Discriminative Domain Adaptation}
  {Adversarial discriminative domain adaptation}.{\BBCQ}
\newblock
\BIn{} \APACrefbtitle {Proceedings of the IEEE Conference on Computer Vision
  and Pattern Recognition (CVPR).} {Proceedings of the ieee conference on
  computer vision and pattern recognition (cvpr).}
\PrintBackRefs{\CurrentBib}

\bibitem [\protect \citeauthoryear {%
M.~Wang%
\ \BBA {} Deng%
}{%
M.~Wang%
\ \BBA {} Deng%
}{%
{\protect \APACyear {2018}}%
}]{%
domadapt_survey}
\APACinsertmetastar {%
domadapt_survey}%
\begin{APACrefauthors}%
Wang, M.%
\BCBT {}\ \BBA {} Deng, W.%
\end{APACrefauthors}%
\unskip\
\newblock
\APACrefYearMonthDay{2018}{}{}.
\newblock
{\BBOQ}\APACrefatitle {Deep visual domain adaptation: A survey} {Deep visual
  domain adaptation: A survey}.{\BBCQ}
\newblock
\APACjournalVolNumPages{Neurocomputing}{312}{}{135-153}.
\newblock
\begin{APACrefURL}
  \url{https://www.sciencedirect.com/science/article/pii/S0925231218306684}
  \end{APACrefURL}
\PrintBackRefs{\CurrentBib}

\bibitem [\protect \citeauthoryear {%
N.~Wang%
\ \protect \BOthers {.}}{%
N.~Wang%
\ \protect \BOthers {.}}{%
{\protect \APACyear {2020}}%
}]{%
wang_autonomous_2020}
\APACinsertmetastar {%
wang_autonomous_2020}%
\begin{APACrefauthors}%
Wang, N.%
, Zhao, X.%
, Zou, Z.%
, Zhao, P.%
\BCBL {}\ \BBA {} Qi, F.%
\end{APACrefauthors}%
\unskip\
\newblock
\APACrefYearMonthDay{2020}{}{}.
\newblock
{\BBOQ}\APACrefatitle {Autonomous damage segmentation and measurement of glazed
  tiles in historic buildings via deep learning} {Autonomous damage
  segmentation and measurement of glazed tiles in historic buildings via deep
  learning}.{\BBCQ}
\newblock
\APACjournalVolNumPages{Computer-Aided Civil and Infrastructure
  Engineering}{35}{3}{277--291}.
\newblock
\begin{APACrefURL} [{2022-07-26}]
  \url{https://onlinelibrary.wiley.com/doi/abs/10.1111/mice.12488}
  \end{APACrefURL}
\PrintBackRefs{\CurrentBib}

\bibitem [\protect \citeauthoryear {%
Weber%
\ \BBA {} Kan{\'{e}}%
}{%
Weber%
\ \BBA {} Kan{\'{e}}%
}{%
{\protect \APACyear {2020}}%
}]{%
detectron}
\APACinsertmetastar {%
detectron}%
\begin{APACrefauthors}%
Weber, E.%
\BCBT {}\ \BBA {} Kan{\'{e}}, H.%
\end{APACrefauthors}%
\unskip\
\newblock
\APACrefYearMonthDay{2020}{}{}.
\newblock
{\BBOQ}\APACrefatitle {Building Disaster Damage Assessment in Satellite Imagery
  with Multi-Temporal Fusion} {Building disaster damage assessment in satellite
  imagery with multi-temporal fusion}.{\BBCQ}
\newblock
\APACjournalVolNumPages{CoRR}{abs/2004.05525}{}{}.
\newblock
\begin{APACrefURL} \url{https://arxiv.org/abs/2004.05525} \end{APACrefURL}
\PrintBackRefs{\CurrentBib}

\bibitem [\protect \citeauthoryear {%
\APACcitebtitle {{{WorldView-2}}}}{%
\APACcitebtitle {{{WorldView-2}}}}{%
{\protect \APACyear {{\protect \bibnodate {}}}}%
}]{%
WorldView2}
\APACinsertmetastar {%
WorldView2}%
\APACrefbtitle {{{WorldView-2}}.} {{{WorldView-2}}.}
\newblock
\APACrefYearMonthDay{{\protect \bibnodate {}}}{}{}.
\newblock
\begin{APACrefURL} [{2022-09-06}]
  \url{https://resources.maxar.com/data-sheets/worldview-2} \end{APACrefURL}
\PrintBackRefs{\CurrentBib}

\bibitem [\protect \citeauthoryear {%
B.~Wu%
\ \protect \BOthers {.}}{%
B.~Wu%
\ \protect \BOthers {.}}{%
{\protect \APACyear {2021}}%
}]{%
Wu_2021_ICCV}
\APACinsertmetastar {%
Wu_2021_ICCV}%
\begin{APACrefauthors}%
Wu, B.%
, Xu, C.%
, Dai, X.%
, Wan, A.%
, Zhang, P.%
, Yan, Z.%
\BDBL {}Vajda, P.%
\end{APACrefauthors}%
\unskip\
\newblock
\APACrefYearMonthDay{2021}{October}{}.
\newblock
{\BBOQ}\APACrefatitle {Visual Transformers: Where Do Transformers Really Belong
  in Vision Models?} {Visual transformers: Where do transformers really belong
  in vision models?}{\BBCQ}
\newblock
\BIn{} \APACrefbtitle {Proceedings of the IEEE/CVF International Conference on
  Computer Vision (ICCV)} {Proceedings of the ieee/cvf international conference
  on computer vision (iccv)}\ (\BPG~599-609).
\PrintBackRefs{\CurrentBib}

\bibitem [\protect \citeauthoryear {%
C.~Wu%
\ \protect \BOthers {.}}{%
C.~Wu%
\ \protect \BOthers {.}}{%
{\protect \APACyear {2021}}%
}]{%
siam-att-unet}
\APACinsertmetastar {%
siam-att-unet}%
\begin{APACrefauthors}%
Wu, C.%
, Zhang, F.%
, Xia, J.%
, Xu, Y.%
\BCBL {}\ \BBA {} Li, G.%
\end{APACrefauthors}%
\unskip\
\newblock
\APACrefYearMonthDay{2021}{}{}.
\newblock
{\BBOQ}\APACrefatitle {Building Damage Detection Using U-Net with Attention
  Mechanism from Pre- and Post-Disaster Remote Sensing Datasets} {Building
  damage detection using u-net with attention mechanism from pre- and
  post-disaster remote sensing datasets}.{\BBCQ}
\newblock
\APACjournalVolNumPages{Remote Sensing}{13}{5}{}.
\newblock
\begin{APACrefURL} \url{https://www.mdpi.com/2072-4292/13/5/905}
  \end{APACrefURL}
\PrintBackRefs{\CurrentBib}

\bibitem [\protect \citeauthoryear {%
R\BHBI T.~Wu%
\ \protect \BOthers {.}}{%
R\BHBI T.~Wu%
\ \protect \BOthers {.}}{%
{\protect \APACyear {2019}}%
}]{%
wuPruningDeepConvolutional2019}
\APACinsertmetastar {%
wuPruningDeepConvolutional2019}%
\begin{APACrefauthors}%
Wu, R\BHBI T.%
, Singla, A.%
, Jahanshahi, M\BPBI R.%
, Bertino, E.%
, Ko, B\BPBI J.%
\BCBL {}\ \BBA {} Verma, D.%
\end{APACrefauthors}%
\unskip\
\newblock
\APACrefYearMonthDay{2019}{}{}.
\newblock
{\BBOQ}\APACrefatitle {Pruning Deep Convolutional Neural Networks for Efficient
  Edge Computing in Condition Assessment of Infrastructures} {Pruning deep
  convolutional neural networks for efficient edge computing in condition
  assessment of infrastructures}.{\BBCQ}
\newblock
\APACjournalVolNumPages{Computer-Aided Civil and Infrastructure
  Engineering}{34}{9}{774--789}.
\newblock
\begin{APACrefURL}
  \url{https://onlinelibrary.wiley.com/doi/abs/10.1111/mice.12449}
  \end{APACrefURL}
\PrintBackRefs{\CurrentBib}

\bibitem [\protect \citeauthoryear {%
Xia%
\ \protect \BOthers {.}}{%
Xia%
\ \protect \BOthers {.}}{%
{\protect \APACyear {2022}}%
}]{%
egctnet}
\APACinsertmetastar {%
egctnet}%
\begin{APACrefauthors}%
Xia, L.%
, Chen, J.%
, Luo, J.%
, Zhang, J.%
, Yang, D.%
\BCBL {}\ \BBA {} Shen, Z.%
\end{APACrefauthors}%
\unskip\
\newblock
\APACrefYearMonthDay{2022}{}{}.
\newblock
{\BBOQ}\APACrefatitle {Building Change Detection Based on an Edge-Guided
  Convolutional Neural Network Combined with a Transformer} {Building change
  detection based on an edge-guided convolutional neural network combined with
  a transformer}.{\BBCQ}
\newblock
\APACjournalVolNumPages{Remote Sensing}{14}{18}{}.
\newblock
\begin{APACrefURL} \url{https://www.mdpi.com/2072-4292/14/18/4524}
  \end{APACrefURL}
\newblock
\begin{APACrefDOI} \doi{10.3390/rs14184524} \end{APACrefDOI}
\PrintBackRefs{\CurrentBib}

\bibitem [\protect \citeauthoryear {%
Xu%
\ \protect \BOthers {.}}{%
Xu%
\ \protect \BOthers {.}}{%
{\protect \APACyear {2021}}%
}]{%
xuRealtimeRegionalSeismic2021}
\APACinsertmetastar {%
xuRealtimeRegionalSeismic2021}%
\begin{APACrefauthors}%
Xu, Y.%
, Lu, X.%
, Cetiner, B.%
\BCBL {}\ \BBA {} Taciroglu, E.%
\end{APACrefauthors}%
\unskip\
\newblock
\APACrefYearMonthDay{2021}{}{}.
\newblock
{\BBOQ}\APACrefatitle {Real-Time Regional Seismic Damage Assessment Framework
  Based on Long Short-Term Memory Neural Network} {Real-time regional seismic
  damage assessment framework based on long short-term memory neural
  network}.{\BBCQ}
\newblock
\APACjournalVolNumPages{Computer-Aided Civil and Infrastructure
  Engineering}{36}{4}{504--521}.
\newblock
\begin{APACrefURL}
  \url{https://onlinelibrary.wiley.com/doi/abs/10.1111/mice.12628}
  \end{APACrefURL}
\PrintBackRefs{\CurrentBib}

\bibitem [\protect \citeauthoryear {%
Zhang%
\ \protect \BOthers {.}}{%
Zhang%
\ \protect \BOthers {.}}{%
{\protect \APACyear {2020}}%
}]{%
zhangConcreteBridgeSurface2020}
\APACinsertmetastar {%
zhangConcreteBridgeSurface2020}%
\begin{APACrefauthors}%
Zhang, C.%
, Chang, C\BHBI c.%
\BCBL {}\ \BBA {} Jamshidi, M.%
\end{APACrefauthors}%
\unskip\
\newblock
\APACrefYearMonthDay{2020}{}{}.
\newblock
{\BBOQ}\APACrefatitle {Concrete Bridge Surface Damage Detection Using a
  Single-Stage Detector} {Concrete bridge surface damage detection using a
  single-stage detector}.{\BBCQ}
\newblock
\APACjournalVolNumPages{Computer-Aided Civil and Infrastructure
  Engineering}{35}{4}{389--409}.
\newblock
\begin{APACrefURL}
  \url{https://onlinelibrary.wiley.com/doi/abs/10.1111/mice.12500}
  \end{APACrefURL}
\PrintBackRefs{\CurrentBib}

\bibitem [\protect \citeauthoryear {%
Zhou%
\ \BBA {} Gong%
}{%
Zhou%
\ \BBA {} Gong%
}{%
{\protect \APACyear {2018}}%
}]{%
zhouAutomatedAnalysisMobile2018}
\APACinsertmetastar {%
zhouAutomatedAnalysisMobile2018}%
\begin{APACrefauthors}%
Zhou, Z.%
\BCBT {}\ \BBA {} Gong, J.%
\end{APACrefauthors}%
\unskip\
\newblock
\APACrefYearMonthDay{2018}{}{}.
\newblock
{\BBOQ}\APACrefatitle {Automated {{Analysis}} of {{Mobile LiDAR Data}} for
  {{Component-Level Damage Assessment}} of {{Building Structures}} during
  {{Large Coastal Storm Events}}} {Automated {{Analysis}} of {{Mobile LiDAR
  Data}} for {{Component-Level Damage Assessment}} of {{Building Structures}}
  during {{Large Coastal Storm Events}}}.{\BBCQ}
\newblock
\APACjournalVolNumPages{Computer-Aided Civil and Infrastructure
  Engineering}{33}{5}{373--392}.
\newblock
\begin{APACrefURL}
  \url{https://onlinelibrary.wiley.com/doi/abs/10.1111/mice.12345}
  \end{APACrefURL}
\PrintBackRefs{\CurrentBib}

\bibitem [\protect \citeauthoryear {%
Zou%
\ \protect \BOthers {.}}{%
Zou%
\ \protect \BOthers {.}}{%
{\protect \APACyear {2022}}%
}]{%
zou_multicategory_2022}
\APACinsertmetastar {%
zou_multicategory_2022}%
\begin{APACrefauthors}%
Zou, D.%
, Zhang, M.%
, Bai, Z.%
, Liu, T.%
, Zhou, A.%
, Wang, X.%
\BDBL {}Zhang, S.%
\end{APACrefauthors}%
\unskip\
\newblock
\APACrefYearMonthDay{2022}{}{}.
\newblock
{\BBOQ}\APACrefatitle {Multicategory damage detection and safety assessment of
  post-earthquake reinforced concrete structures using deep learning}
  {Multicategory damage detection and safety assessment of post-earthquake
  reinforced concrete structures using deep learning}.{\BBCQ}
\newblock
\APACjournalVolNumPages{Computer-Aided Civil and Infrastructure
  Engineering}{37}{9}{1188--1204}.
\newblock
\begin{APACrefURL} [{2022-07-26}]
  \url{https://onlinelibrary.wiley.com/doi/abs/10.1111/mice.12815}
  \end{APACrefURL}
\PrintBackRefs{\CurrentBib}

\end{thebibliography}

\end{document}